\RequirePackage{snapshot}
\documentclass{article} 
\usepackage{iclr2020_conference,times}

\usepackage{amsmath,amsfonts,bm}

\def\eqref#1{equation~\ref{#1}}

\def\1{\bm{1}}

\DeclareMathAlphabet{\mathsfit}{\encodingdefault}{\sfdefault}{m}{sl}
\SetMathAlphabet{\mathsfit}{bold}{\encodingdefault}{\sfdefault}{bx}{n}
\newcommand{\tens}[1]{\bm{\mathsfit{#1}}}

\newcommand{\etens}[1]{\mathsfit{#1}}

\usepackage{url}
\usepackage{tabularx}
\usepackage{booktabs}       
\usepackage{amsfonts}       
\usepackage{nicefrac}       
\usepackage{microtype}      
\usepackage{graphicx}
\usepackage[font=small]{caption}
\usepackage{color}
\usepackage{xcolor,colortbl}  
\usepackage{amsmath}
\usepackage{algorithm}
\usepackage{algorithmic}
\usepackage{multirow}
\usepackage{array}

\usepackage[pagebackref=true,breaklinks=true,letterpaper=true,colorlinks,bookmarks=false]{hyperref}

\usepackage[pagebackref=true,breaklinks=true,letterpaper=true,colorlinks,bookmarks=false]{hyperref}

\definecolor{Gray}{gray}{0.9}
\newcolumntype{g}{>{\columncolor{Gray}}c}  
\newcolumntype{H}{>{\setbox0=\hbox\bgroup}c<{\egroup}@{}}  
\definecolor{GrayLine}{gray}{0.7}
\newcolumntype{Y}{>{\centering\arraybackslash}X}
\newcommand{\method}[0]{MetaPix}
\newcommand{\poseim}[0]{Pose2Im}
\newcommand{\posewarp}[0]{Posewarp}

\title{MetaPix: Few-Shot Video Retargeting}

\author{Jessica Lee \quad Deva Ramanan \quad Rohit Girdhar \\ 
Carnegie Mellon University\\
{\small \url{https://imjal.github.io/MetaPix}}
}

\iclrfinalcopy 
\begin{document}

\maketitle

\begin{abstract}
    We address the task of retargeting of human actions from
    one video to another. We consider the challenging setting where only a few frames of the target is available.
    The core of our approach is a conditional generative 
    model that can transcode input skeletal poses (automatically extracted with an off-the-shelf pose estimator) to output target frames. However, it is challenging to build a universal transcoder because humans can appear wildly different due to clothing and background scene geometry. Instead, we learn to adapt -- or {\em personalize} -- a universal generator to the particular human and background in the target. To do so, we make use of {\em meta}-learning to discover effective strategies for on-the-fly personalization. One significant benefit of meta-learning is that the personalized 
    transcoder naturally enforces temporal coherence across its generated frames; all frames contain consistent clothing and background geometry of the target.
    We experiment on in-the-wild internet
    videos and images and show our approach improves over widely-used baselines for
    the task. 
\end{abstract} \section{Introduction}

One of the hallmarks of human intelligence is the ability to imagine. For example, given an image of a never-before-seen person, one can easily imagine them performing different actions. To do so, we make use of years of experience watching humans act and interact with the world. We implicitly encode the rules
of physical transformations of humans, objects, clothing and so on. Crucially, we effortlessly adapt or {\em retarget} those universal rules to a specific human and environment - a child on a playground will likely move differently than an adult walking into work. 
Our goal in this work is to develop models that similarly learn
to generate human motions by specializing universal knowledge to a particular target human and target environment, given only a few samples of the target. 

It is attractive to tackle such video generation tasks using the framework of generative (adversarial) neural networks (GANs). Past work has cast the core computational problem as one of conditional image generation where input source poses (automatically extracted with an off-the-shelf pose estimator) are transcoded into image frames ~\citep{balakrishnan2018synthesizing,siarohin2018deformable,ma2017pose}. However, it is notoriously challenging to build generative models that are capable of synthesizing diverse, in-the-wild imagery. Notable exceptions make use of massively-large networks trained on large-scale compute infrastructure~\citep{brock2018large}. However, modestly-sized generative networks perform quite well at synthesis of targeted domains (such as faces~\citep{Recycle-GAN} or facades~\citep{pix2pix2016}).
A particularly successful approach to generating from pose-to-image is training of specialized -- or {\em personalized} -- models to particular scenes. These often require large-scale target datasets, such as 20 minutes of footage in a target lab setting~\citep{chan2018everybody} 

The above approaches make use of personalization as an implicit but crucial ingredient, by {\em on-the-fly} training of a generative model tuned to the particular target domain of interest. Often, personalization is operationalized by fine-tuning a generic model on the specific target frames of interest. Our key insight is recasting personalization as an {\em explicit} component of a video-retargeting engine, allowing us to make use of meta-learning to {\em learn} how best to fine-tune (or personalize) a generic model to a particular target domain. We demonstrate that (meta)learning-to-fine-tune is particularly effective in the few-shot regime, where few target frames are available. From a technical perspective, one of our contributions is extending meta-learning to GANs, which is nontrivial because both a generator and discriminator need to be adversarially fine-tuned.

To that end, we propose \method{}, a novel approach to personalization for video retargeting.
Our formulation treats personalization as a few-shot learning problem, where the 
task is to {\em adapt} a generic generative model of human actions to a specific 
person given a few samples of their appearance. Our formulation is agnostic
to the actual generative model used, and is compatible with both
pose-conditioned transfer~\citep{balakrishnan2018synthesizing}
or generative~\citep{chan2018everybody} approaches.
Taking inspiration from the recent successes of meta-learning approaches
for few-shot tasks~\citep{nichol2018first,finn2017model}, 
we propose a novel formulation by 
adapting the popular first-order meta-learning algorithm Reptile~\citep{nichol2018first}
for jointly learning initial weights for both the generator and discriminator.
Hence, our model is optimized for efficient adaptation (personalization),
given only a few samples and on a computational budget, and obtains stronger performance
compared to a model not optimized in this form. Interestingly, we find this personalized
model naturally enforces strong temporal coherence in the generated frames, even though it is not
explicitly optimized for that task.

\begin{figure}[t]
    \centering
    \includegraphics[width=\linewidth]{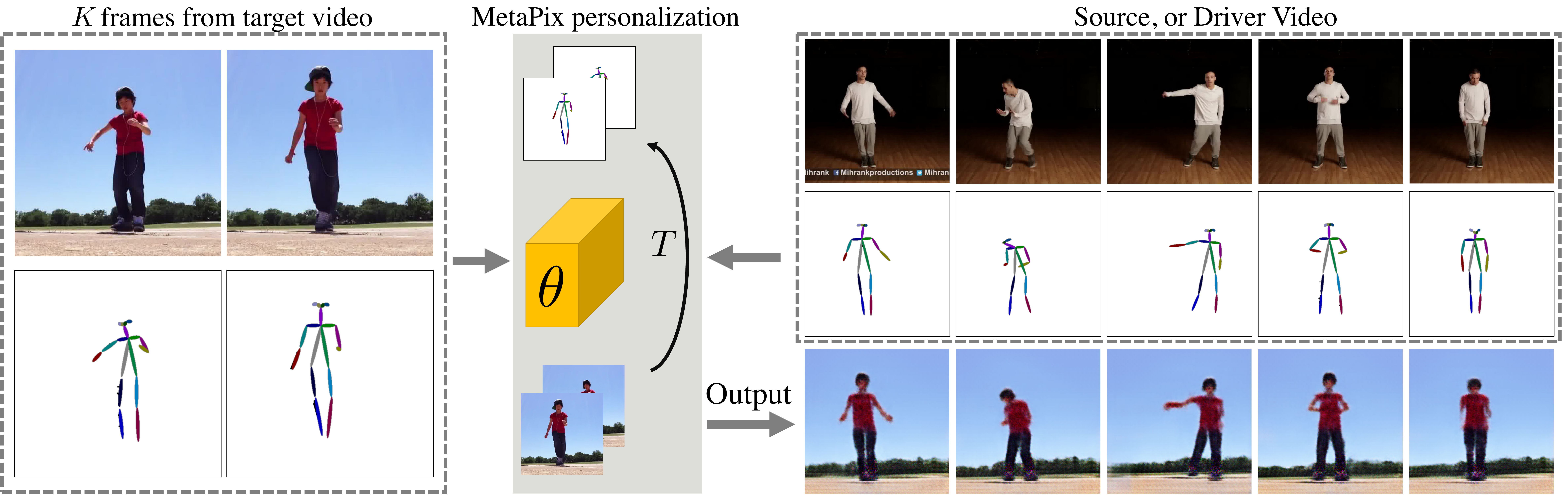}
    \caption{
    {\bf Video retargeting on a budget.}
    Our goal is to retarget a source video to a target, 
    {\em quickly} (running a few, $T$, iterations for adaptation) and
    {\em efficiently} (given a few, $K$, frames from the target domain).
    We achieve that by (meta)learning a model $\theta$ that is 
    able to quickly and efficiently adapt to a given target video.
    }\label{fig:teaser}
\end{figure}
 \section{Related Work}

{\noindent \bf Deep generative modeling.}
There has been a growing interest in using deep networks for generative modeling of visual data,
particularly images. Popular techniques include
Variational Auto-Encoders (VAEs)~\citep{kingma2013auto}
and Generative Adversarial Networks (GANs)~\citep{goodfellow2014generative}. 
Particularly, GAN based techniques have shown strong performance for 
various tasks
such as conditional image generation~\citep{brock2018large},
image-to-image translation~\citep{pix2pix2016,wang2018pix2pixHD,CycleGAN2017,balakrishnan2018synthesizing},
unsupervised translation~\citep{CycleGAN2017}
and domain adaptation~\citep{Hoffman_cycada2017}.
More recently, these techniques have been extended to video tasks,
such as generation~\citep{vondrick2016generating},
future prediction~\citep{finn2016unsupervised} and
translation~\citep{Recycle-GAN,wang2018vid2vid}.
Our work explores generative modeling from a few samples,
with our main focus being the task of video translation.
There has been some prior work in this direction~\citep{zakharov2019few}, though is largely limited to faces and portrait images.

{\noindent \bf Motion transfer and video retargeting.}
This refers to the task of driving a video of a person or 
a cartoon character given another video~\citep{gleicher1998retargetting}.
While there exist
some unsupervised techniques~\citep{Recycle-GAN} to do so,
most successful approaches for articulated bodies involve using pose 
as an intermediate supervision.
Recently, there have been two broad categories of approaches 
that have been employed for this task: 1) Learning to transform 
an image into another, given pose as
input, either in 2D~\citep{zhou2019dance,balakrishnan2018synthesizing,siarohin2018deformable,ma2017pose}
or 3D~\citep{liu2018neural,neverova2018dense,walker2017pose}.
And 2) Learning a model to directly generate images given a pose 
as input (or, {\em \poseim{}})~\citep{chan2018everybody}. 
The former approaches tend to be more sophisticated, separately 
generating foreground and background pixels, and tend to perform 
slightly better than the latter. However, they typically 
learn a generic model across datasets that can
transfer from a single frame, whereas the latter 
can learn a more holistic reconstruction by learning 
a specific model for a video. 
Our approach is complementary to
such transfer approaches, and be applied on top of either, as 
we discuss in Section~\ref{sec:approach}.

{\noindent \bf Few-shot learning.}
Low shot learning paradigms attempt to learn a model using 
very small amount of training data~\citep{thrun1996learning},
typically for visual recognition tasks. 
Classical approaches build generative models that 
share priors across the various 
categories~\citep{fei2006one,salakhutdinov2012one}.
Another category of approaches attempt to learn feature 
representations invariant to intra-class
variations by using hallucinated data~\citep{hariharan2017low,wang2018low} 
or specialized
training procedures/loss 
functions~\citep{wang2016learning,bart2005cross}.
More recently, it has been framed as a `learning-to-learn' or a 
meta-learning problem. The key idea is to directly optimize the 
model, for the eventual few-shot adaptation task, where the model 
is finetuned using a few examples~\citep{finn2017model}.
Alternatively, it has also been explored in form of 
directly predicting classifier
weights~\citep{bertinetto2016learning,wang2016learning_to_learn,wang2017learning_model_tail,misra2017red}.

{\noindent \bf Meta Learning.}
The goal of metalearning is to learn models that are good at learning,
similar to how humans are able to quickly and efficiently 
learn to do a new task. 
Many different approaches have been explored to that end.
One direction involves learning weights through recurrent
networks like LSTM~\citep{hochreiter2001learning,santoro2016meta,duan2016rl}.
More commonly, meta-learning has been used as a way to learn an 
initialization for a network, that is finetuned at {\em test time}
on a new task. A popular approach in this direction 
is MAML~\citep{finn2017model}, where the parameters are directly optimized for the 
test time performance of the task it needs to adapt to. This is performed by 
backpropagating through the finetuning process by computing second order gradients.
They and others~\citep{andrychowicz2016learning} have also proposed first-order methods like FOMAML that
forego the need to compute second order gradients, making it 
more efficient at empirically small drop in performance. 
However, most of these works still tend to have the requirement of SGD to be used 
as the task optimizer. A recently proposed meta-learning algorithm, Reptile~\citep{nichol2018first},
 forgoes that constraint by proposing a much simpler first order 
meta learning algorithm that is compatible with 
any black box optimizer.
 \section{Our Approach}\label{sec:approach}
We now describe \method{} in detail. 
To reiterate, our goal is to learn a generic model of human motion,
parameterized by $\theta$, that can 
quickly and efficiently be {\em personalized}
for a specific person.
We define speed and efficiency requirements in terms of two parameters:
computation/iterations ($T$) and the number samples 
required for personalization ($K$), respectively.
We now describe the base architecture, \method{} training setup, and the
implementation details.

\begin{figure}[t]
    \centering
    \includegraphics[width=\linewidth]{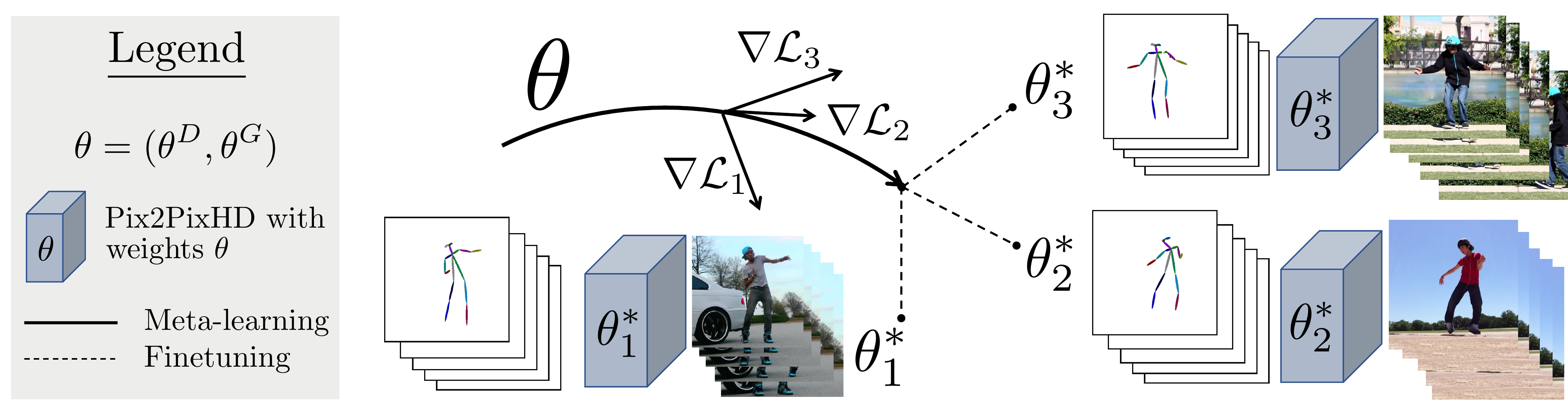}
    \caption{
    {\bf Meta-learning for video retargeting.}
    Our goal is to learn a generic retargeting model $\theta$
    (parameters of a Pix2PixHD~\citep{wang2018pix2pixHD} in our case),
    such that it can be efficiently 
    personalized for a specific person given only a few samples/frames of their 
    appearance. We achieve it using meta-learning, where our model is optimized
    for personalization to a new person, given only $K$ samples of their appearance,
    and being trained for $T$ iterations.
    }\label{fig:metalearning}
\end{figure}

\paragraph{Base retargeting architecture.}
We build upon popular video retargeting architectures. Notably, there are two 
common approaches in literature:1) Learning a transformation from one image to another,
conditioned on the pose~\citep{zhou2019dance,balakrishnan2018synthesizing} and 2) Learning a mapping from pose to RGB (\poseim{}), like~\citep{chan2018everybody}.
Both obtain strong performance and amenable 
to the speed and efficiency constraints we are interested in.
For example in $K$-shot setting (i.e.\ to learn a model using $K$ frames),
one can train the \poseim{} mapping using the $K$ frames in the former case,
or use the $C^K_2$ pairs from $K$ frames to learn a transformation function from one of the $K$ images to another in the latter
case. They are also both compatible with our \method{} optimization discussed next. 

\poseim{}~\citep{chan2018everybody} approaches essentially build upon
image-to-image translation methods~\citep{pix2pix2016,wang2018pix2pixHD},
where the input is a rendering of the body joints, and the output is an RGB image.
The model consists of an encoder-decoder style generator $G$. It is trained using a
combination of perceptual reconstruction losses~\citep{johnson2016perceptual},
implemented using an $L_1$ penalty
over VGG~\citep{Simonyan_14a} features and discriminator losses, where we 
train a separate discriminator network $D$ that is trained to differentiate the 
generated images from real images.
The reconstruction loss forces it to be close to 
the ground truth, potentially leading to blurry outputs.
Adding the discriminator helps fix that, as it forces the output onto the 
manifold of real images.
Given its strong performance, we use Pix2PixHD~\citep{wang2018pix2pixHD}
as our base architecture for \poseim{}.
For brevity, we skip a complete description of the model architecture, and refer the reader
to~\citep{wang2018pix2pixHD} for more details.

Pose Transfer~\citep{balakrishnan2018synthesizing,zhou2019dance},
on the other hand, takes a source image of a person and a target pose, and generates an image of the source person in that target pose. These approaches typically segment the limbs, transform their position
as in the target pose, and generate the target image by combining the transformed limbs and segmented background by using a generative network like a U-Net~\citep{unet}. 
These approaches can leverage learning to move pixels instead of having to generate color and background image from a learned representation.
We utilize the \posewarp{} method~\citep{balakrishnan2018synthesizing} as our base Pose Transfer architecture due to available implementation.

\paragraph{\method{}.}

\method{} builds upon the base retargeting architecture by optimizing
it for few-shot and fast adaptation for personalization. 
We achieve that by taking inspiration from the literature on few-shot learning, where 
meta-learning has shown promising results.
We use a recently introduced first-order meta-learning technique, 
Reptile~\citep{nichol2018first}. As compared to 
the more popular technique, MAML~\citep{finn2017model},
it is more efficient as it does not compute a second gradient and is amenable 
to work with arbitrary optimizers as it does not need to backpropagate through the 
optimization process. Given that GAN architectures are hard to optimize, 
Reptile suits our purposes of its ability to use Adam~\citep{kingma2015adam},
the default optimizer for Pix2PixHD, as our task optimizer.
Figure~\ref{fig:metalearning} illustrates the high level idea of our approach,
which we describe in detail next.

We start with either a \poseim{} or a Pose Transfer trained base model.
We then finetune this model as described in Algorithm~\ref{alg:reptile}.
Note that Pix2PixHD is based on a GAN, so has two network weights to be 
optimized, the generator ($\theta_G$) and discriminator ($\theta_D$).
In each meta-iteration, we sample a {\em task}: in our case a set of $K$ frames from a 
new video to personalize to. We then finetune the current model parameter to that 
video over $T$ iterations, and update the model parameters in the direction
of the personalized parameters using a meta learning rate $\epsilon$.
We optimize both $\theta_D$ and $\theta_G$ jointly at each step.
Note that \posewarp{} employs a more complicated two-stage training procedure, and we metalearn only the first stage (which has no discriminator) for simplicity.

\begin{algorithm}[t]
  \begin{algorithmic} 
    \STATE Initialize $\theta_{D}, \theta_{G}$ from pretrained weights
    \FOR{iteration = $1, 2, ...$}
      \STATE Sample $K$ pose image pairs from the same shot randomly
      \STATE Compute $\widetilde{\theta_{D}}, \widetilde{\theta_{G}} = \text{Pix2PixHD}_{K}^{T}(\theta_{D}, \theta_{G})$, for $K$ images and $T$ iterations
      \STATE Update $\theta_{D} = \theta_{D} - \epsilon(\widetilde{\theta_{D}} - \theta_{D})$
      \STATE Update $\theta_{G} = \theta_{G} - \epsilon(\widetilde{\theta_{G}} - \theta_{G})$
    \ENDFOR
  \end{algorithmic}
    \caption{Meta-learning for video re-targeting for the \poseim{} setup.}\label{alg:reptile}
\end{algorithm}

\paragraph{Implementation Details.}

We implement \method{} for the \poseim{} base model by building upon a public Pix2PixHD
implementation\footnote{\url{https://github.com/NVIDIA/pix2pixHD/}} in PyTorch, and perform 
all experiments on a 4 TITAN-X or GTX 1080Ti GPU node.
We follow the hyperparameter setup as proposed in~\citep{wang2018pix2pixHD}.
We represent the pose using a multi-channel heatmap image,
and input and output are $512 \times 512$px RGB images.
The generator consists of $16$ 
convolutional and deconvolutional layers, and is trained
with a equally weighted combination of GAN, Feature Matching, and VGG losses. 
Initially, we pretrain the model on a large corpus of videos to learn a generic \poseim{} model
as described in Section~\ref{sec:exp}.
During this pretraining stage, the model is trained on all of the training
frames for 10 epochs using learning rate of 0.0002
and batch size of 8 distributed over the 4 GPUs.
We experimented with multiple learning rates including $0.2, 0.02, 0.002$; however, we observed that higher
learning rates caused the training to diverge.
When finetuning for personalization, given $K$ frames and 
a computational budget $T$, 
we train the first $\frac{T}{2}$ iterations using a
constant learning rate of 0.0002,
and the remaining iterations using a linear decay to 0, following~\citep{wang2018pix2pixHD}.
The batch size is fixed to 8, and for $K < 8$, we repeat the frames to
get 8 images for the batch. 
For the metalearning, we set the meta learning rate, $\epsilon=1$ with a linear decay to 0,
and train 300 meta-iterations. We also experiment with meta learning rate, $\epsilon=0.1$, however, 
was much slower to converge.
To potentially stabilize metatraining, we experiment with differing numbers of updates to the generator and discriminator during iterations of Alg.~\ref{alg:reptile}, as well as simplified objective functions.
Recall that the GAN loss adds significant complexity due to the presence of a discriminator that need also be adversarially finetuned. 
In total, our metalearning takes 1 day of training time on 4 GPUs.
For the Pose Transfer base model, we apply \method{} in a similar fashion on top of \posewarp{}\footnote{\url{https://github.com/balakg/posewarp-cvpr2018}}, 
using the author provided pretrained weights.
We will release the \method{} source code for details.  \section{Experiments}\label{sec:exp}

We now experimentally evaluate \method{}.
We start by describing the datasets used and evaluation metrics.
We then describe our base \poseim{} and Pose Transfer setup,
followed by training that model using \method{}.
Finally, we analyze and ablate the various design choices in \method{}.

\begin{figure}[t]
    \centering
    \includegraphics[height=1in,width=\linewidth]{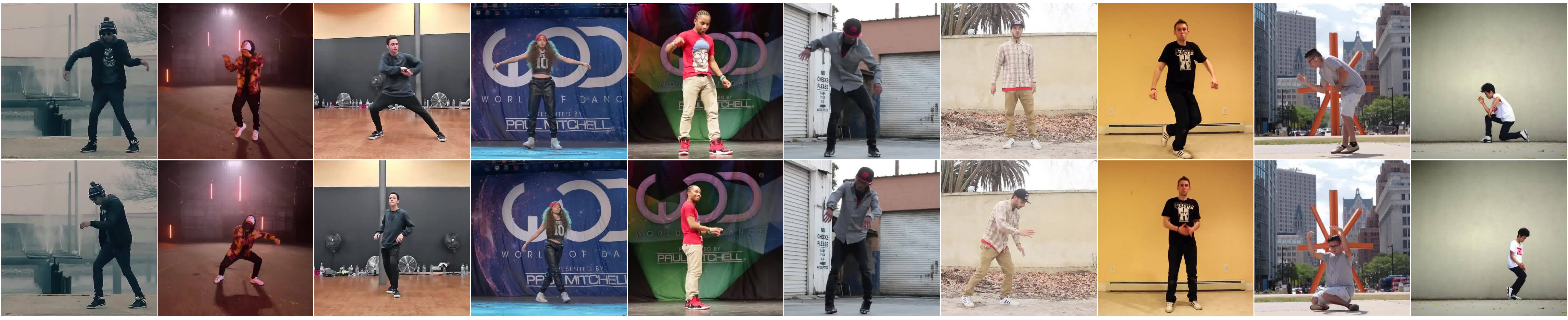}
    \caption{
    {\bf Training data.}
    Frames from the additional training data we collected. We download 
    10 videos from YouTube, distinct from the ones used for 
    personalization and evaluation.
    }\label{fig:train_data}
\end{figure}

\vspace{-0.1in}
\subsection{Datasets and Evaluation}

We train and evaluate our approach on in-the-wild internet videos. 
Due to the lack of a standard benchmark for such retargeting tasks,
we use the dataset as described in~\citep{zhou2019dance} as our test set.
These are a set of 8 videos downloaded from youtube, each 4-12 minutes long.
We refer the reader to Figure 1 in~\citep{zhou2019dance} for sample
frames from this dataset.
Additionally, we collect a set of 10 more dance videos from YouTube
(distinct from the above 8), as our pre-training and meta-learning corpus. We provide the list of YouTube video IDs for both in the supplementary.
Our models are only trained on these videos, and videos from~\citep{zhou2019dance}
are only used for personalization (using $K$ frames) and evaluation.
Figure~\ref{fig:train_data} shows sample frames from these newly collected videos.

{\noindent \bf Evaluation and Metrics:}
Similar to~\citep{zhou2019dance}, we split each of the 8 test videos into a training and
test sequence in 0.85:0.15 ratio, and sample $K$ training and 2000 test
frames from the test sequence. 
We use the same metrics as in~\citep{zhou2019dance} for ease of comparison:
Mean Squared Error (MSE), Structured Similarity Index (SSIM)
and Peak Signal-to-Noise Ratio (PSNR). Each of these are averaged over the 
2000 test frames from each of the 8 test videos. 
To compare our baselines and our method, pose retargeting as a task aims to minimize MSE and maximize SSIM and PSNR. 

\vspace{-0.1in}
\subsection{Evaluating \method{}}

\begin{table}[t]
    \centering
    \small
    \caption{
        {\bf Adding \method{}.}
        We compare the Pix2PixHD and \posewarp{} models' performance.
        The models are initialized either at random, pretrained on the entire training set of videos,
        or meta-learned with \method{}.
        Constraining the amount of frames and compute used 
        at test time leads to a drop in performance as expected. In general, 
        Pix2PixHD performs well with large training sets because it can learn to model
        subtle dependancies between pose and appearance. In contrast, \posewarp{} performs
        better with smaller training sets because it directly transfers pixels from the 
        background and foreground. However, in both cases, meta-learning with \method{} produces better performance, given the exact 
        same test time constraints.
        We compare these methods
        qualitatively in Figure~\ref{fig:improve}.
        }\label{tab:baseline}
    \begin{tabular}{lccccccc}
    \toprule
    Method & Init & $K$ & $T$ &  SSIM & PSNR & MSE \\
    \midrule
    \multirow{5}{*}{Pix2PixHD} & Random & $\infty$ & $\infty$ & $0.68$ &  $19.56$ & $2,427.18$ \\ 
     & Pretrain & $\infty$ & $\infty$ & $0.69$ & $19.31$ & $2,673.40$\\  
     \arrayrulecolor{GrayLine}
     \cmidrule(){2-7} 
     & Random & $5$ & $20$  & $0.08$ & $9.51$ & $7,801.19$\\
     & Pretrain & $5$ & $20$  & $0.35$ & $12.00$ & $5,576.28$\\
     & \method{} & $5$ & $20$ & $0.39$ & $13.73$ &	$4,696.00$  \\
    \arrayrulecolor{GrayLine}
    \midrule
    \multirow{4}{*}{Posewarp} & Pretrain & $\infty$ & $\infty$ & $0.58$ & $17.51$ &  $2,901.13$ \\  
    \cmidrule(){2-7}
    & Random & $5$ & $20$  & $0.40$ & $12.25$ & $4,670.81$\\
     & Pretrain & $5$ & $20$  & $0.55$ & $16.53$ & $3,140.91$ \\
     & \method{} & $5$ & $20$ & $0.56$ & $16.94$ &	$2,962.67$  \\
    \arrayrulecolor{black}
    \bottomrule
    \end{tabular}
\end{table}

\begin{figure}
    \centering
    \setlength{\tabcolsep}{2pt}
    \resizebox{\linewidth}{!}{
    \begin{tabular}{cccccc}
    Ground Truth & Pix2PixHD $K=\infty$ & Pretrained Pix2PixHD &  \method{} Pix2PixHD & Pretrained Posewarp & \method{} Posewarp \\
    \includegraphics[width=0.25\linewidth]{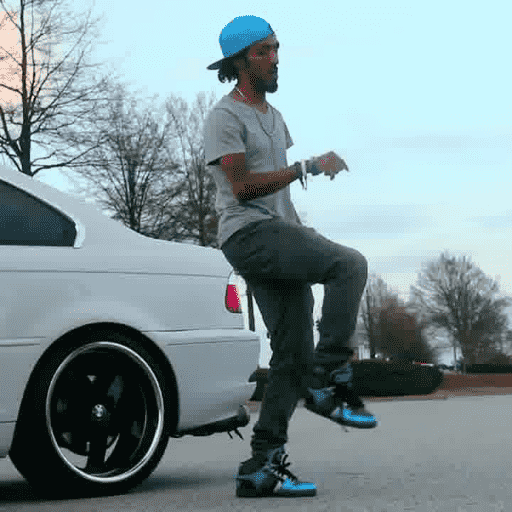} &
    \includegraphics[width=0.25\linewidth]{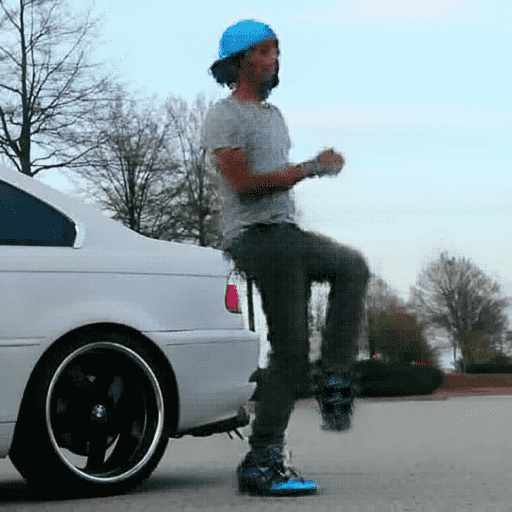} & 
    \includegraphics[width=0.25\linewidth]{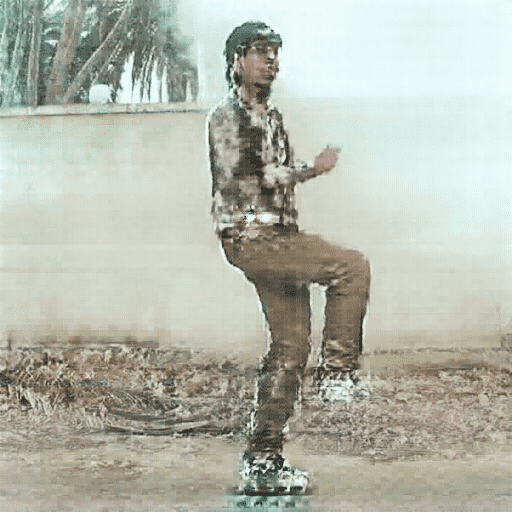} & 
    \includegraphics[width=0.25\linewidth]{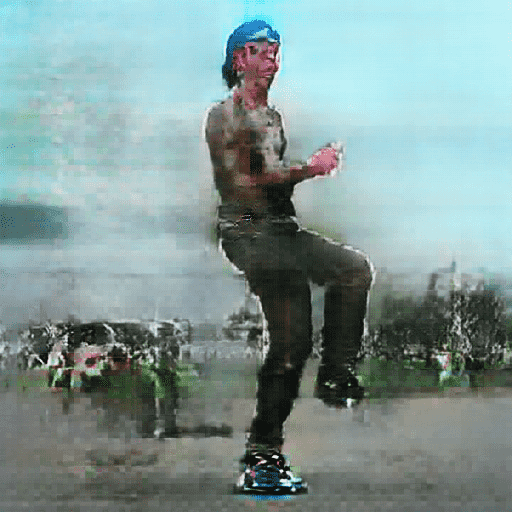} &
    \includegraphics[width=0.25\linewidth]{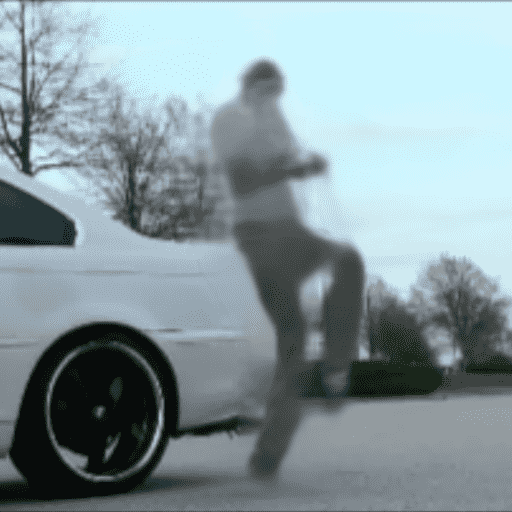} &
    \includegraphics[width=0.25\linewidth]{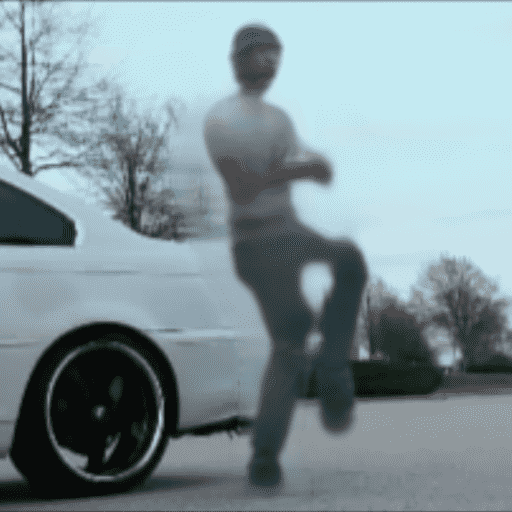} \\
    \includegraphics[width=0.25\linewidth]{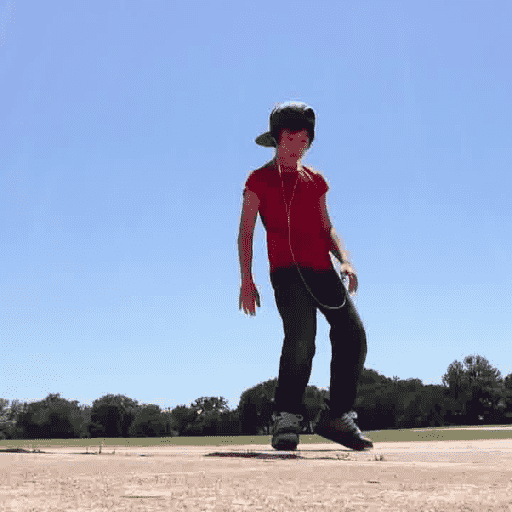} &
    \includegraphics[width=0.25\linewidth]{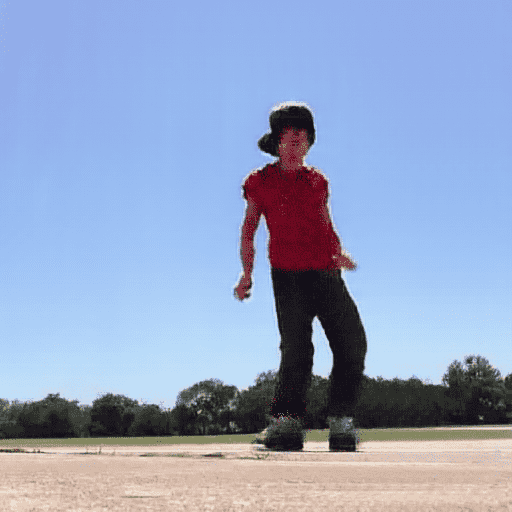} & 
    \includegraphics[width=0.25\linewidth]{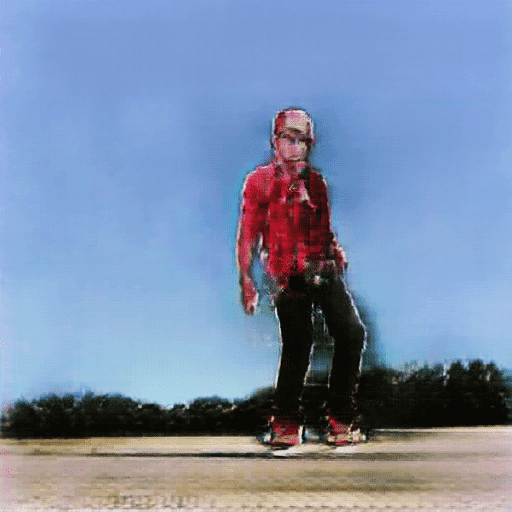} & 
    \includegraphics[width=0.25\linewidth]{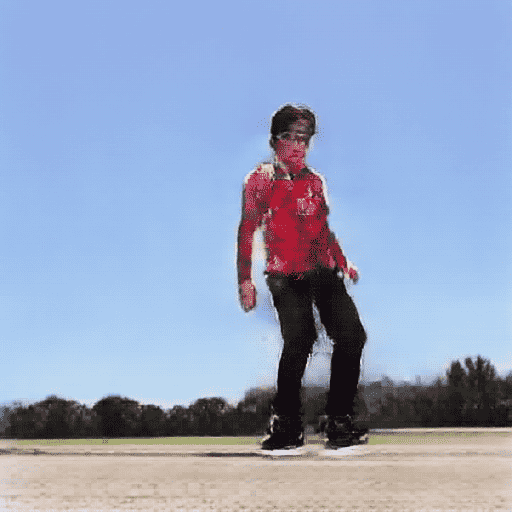} &
    \includegraphics[width=0.25\linewidth]{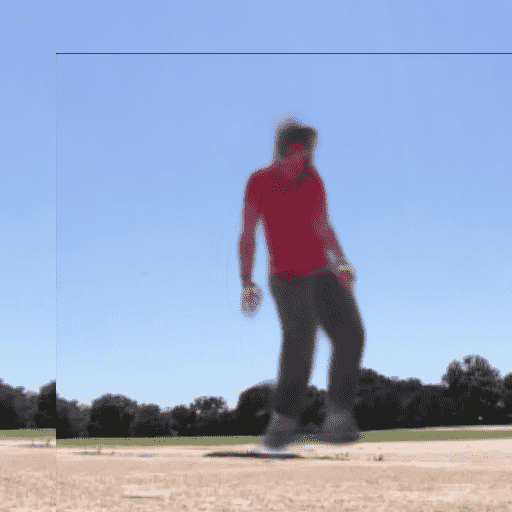} &
    \includegraphics[width=0.25\linewidth]{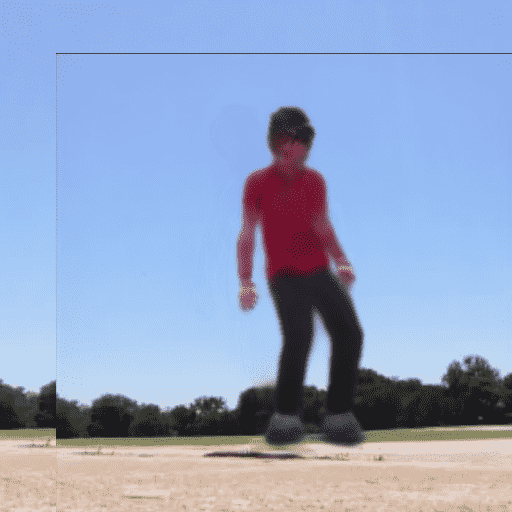} \\
    \includegraphics[width=0.25\linewidth]{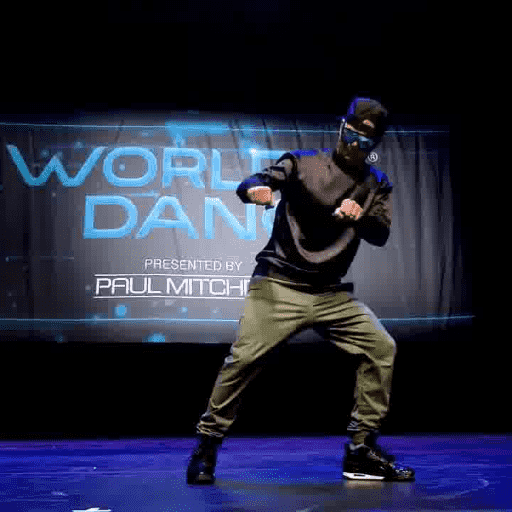} &
    \includegraphics[width=0.25\linewidth]{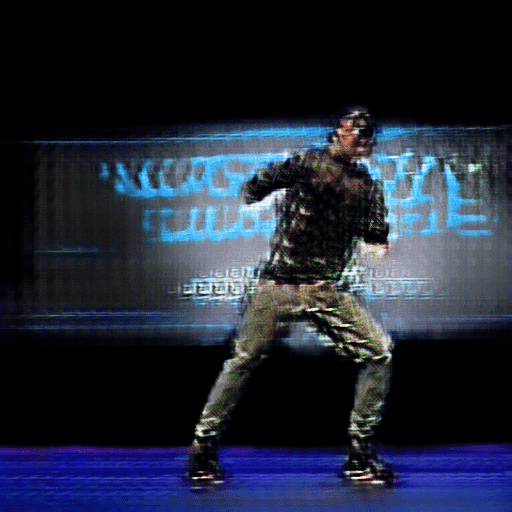} & 
    \includegraphics[width=0.25\linewidth]{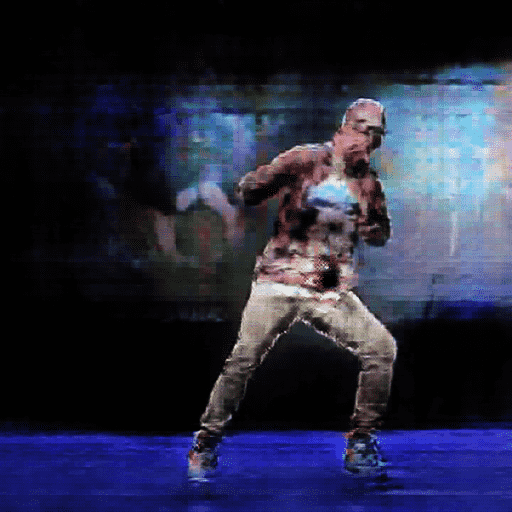} & 
    \includegraphics[width=0.25\linewidth]{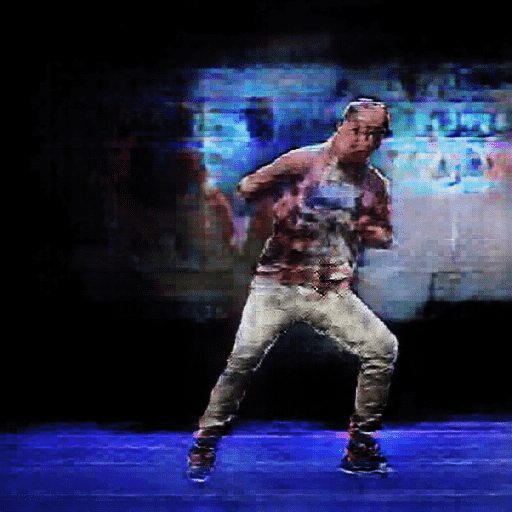} &
    \includegraphics[width=0.25\linewidth]{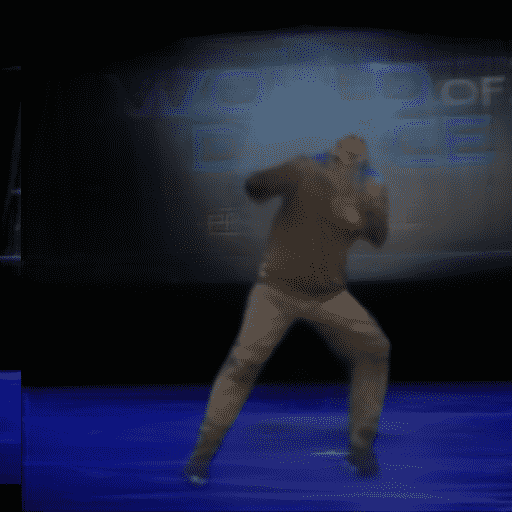} &
    \includegraphics[width=0.25\linewidth]{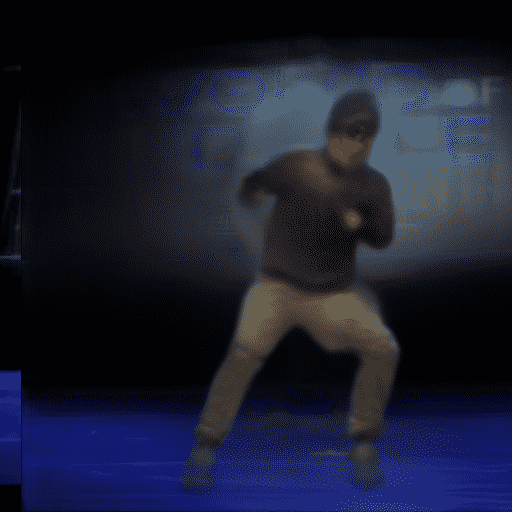} \\
    \includegraphics[width=0.25\linewidth]{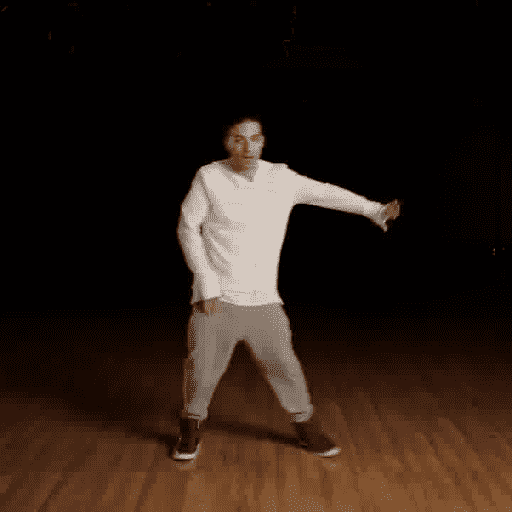} &
    \includegraphics[width=0.25\linewidth]{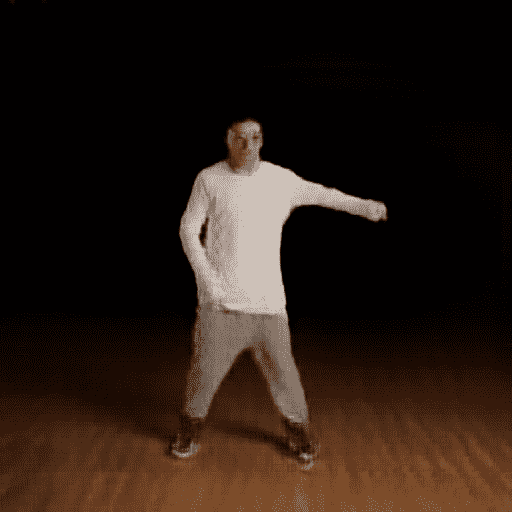} & 
    \includegraphics[width=0.25\linewidth]{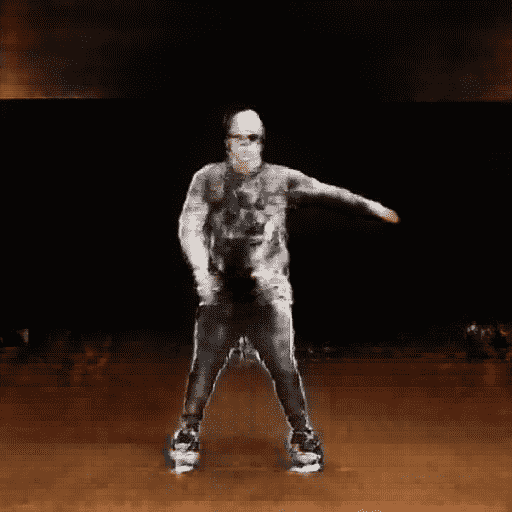} & 
    \includegraphics[width=0.25\linewidth]{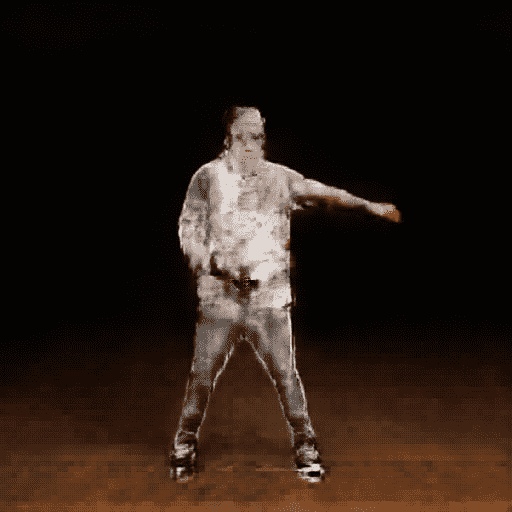} &
    \includegraphics[width=0.25\linewidth]{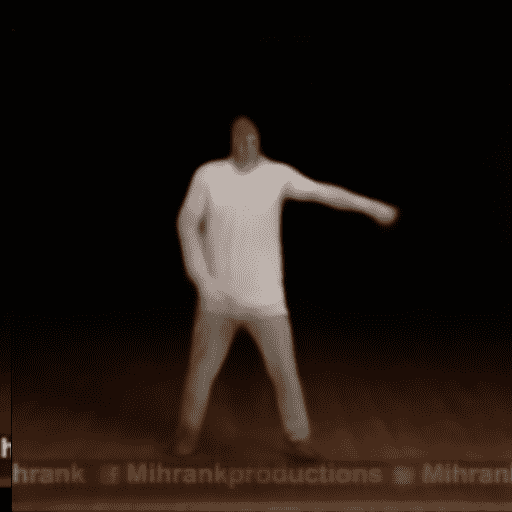} &
    \includegraphics[width=0.25\linewidth]{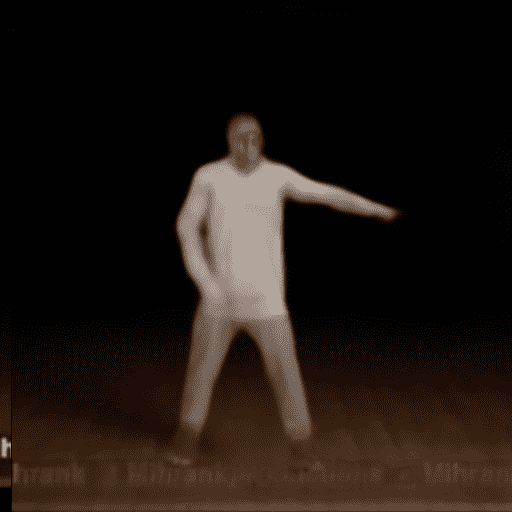} \\
    \includegraphics[width=0.25\linewidth]{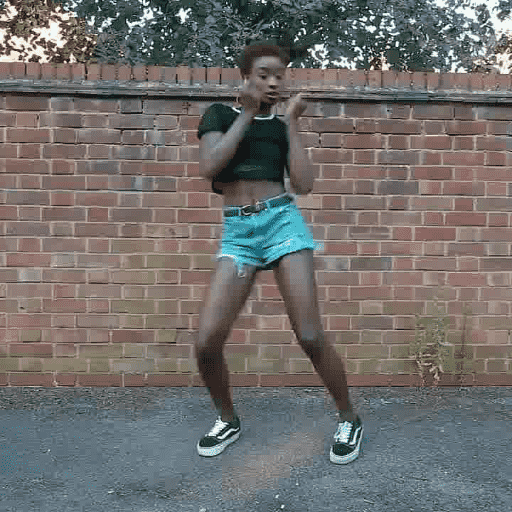} &
    \includegraphics[width=0.25\linewidth]{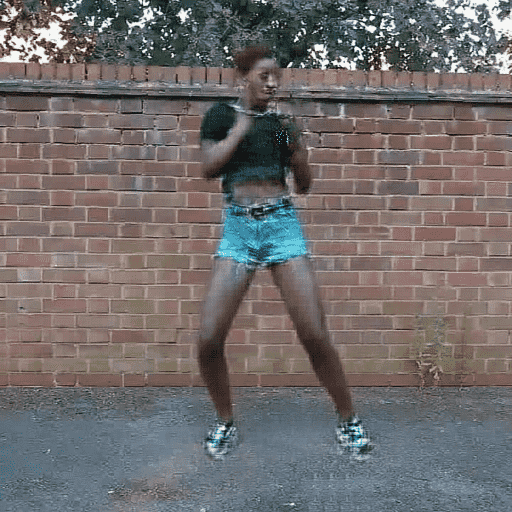} & 
    \includegraphics[width=0.25\linewidth]{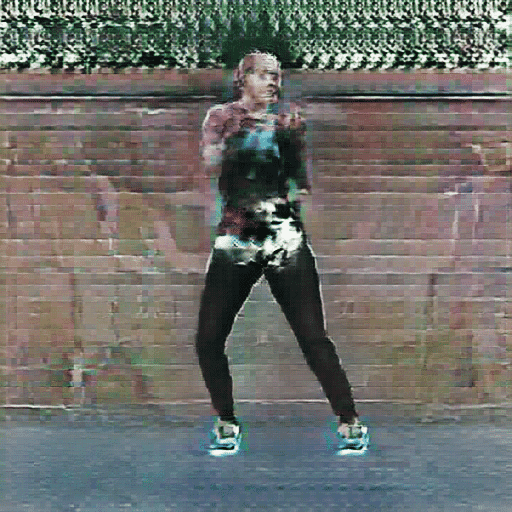} & 
    \includegraphics[width=0.25\linewidth]{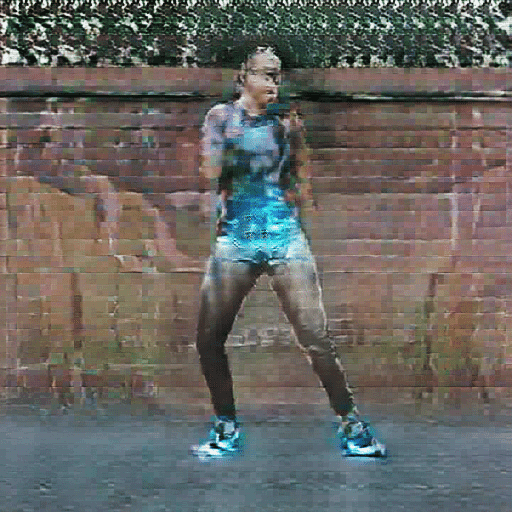} &
    \includegraphics[width=0.25\linewidth]{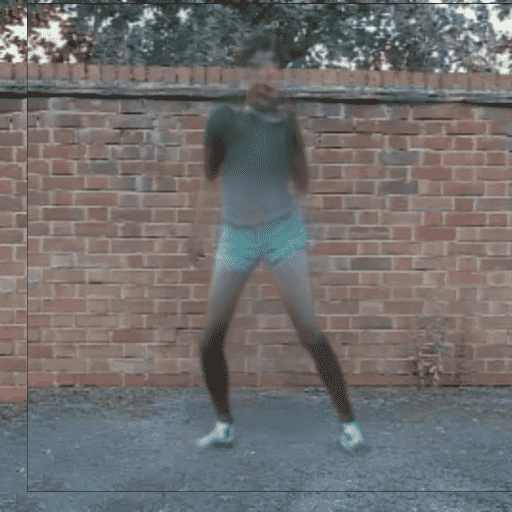} &
    \includegraphics[width=0.25\linewidth]{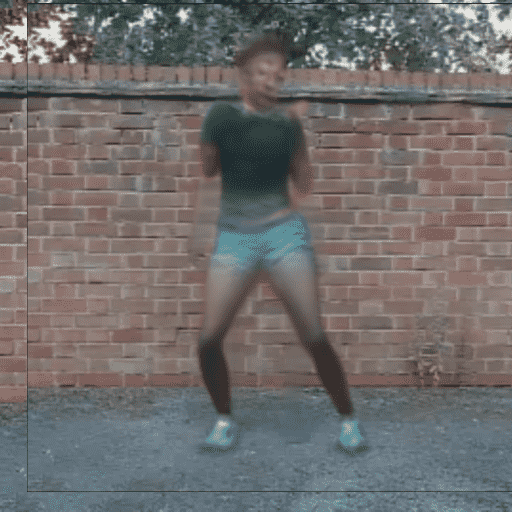} \\
    \includegraphics[width=0.25\linewidth]{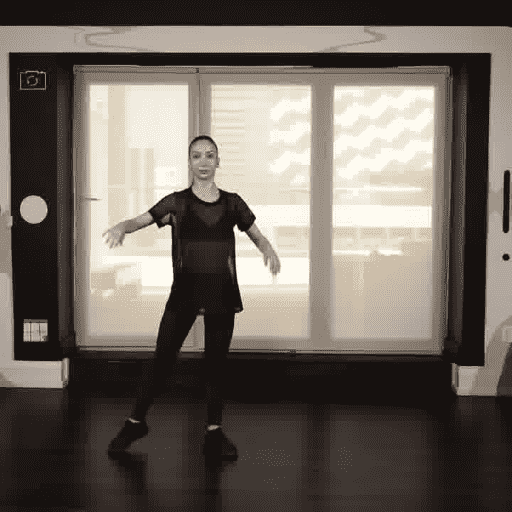} &
    \includegraphics[width=0.25\linewidth]{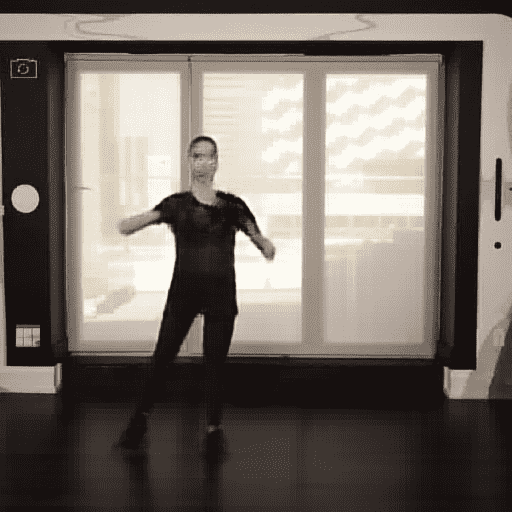} & 
    \includegraphics[width=0.25\linewidth]{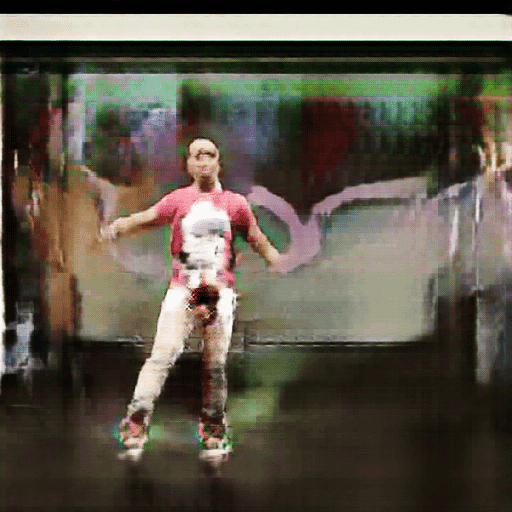} & 
    \includegraphics[width=0.25\linewidth]{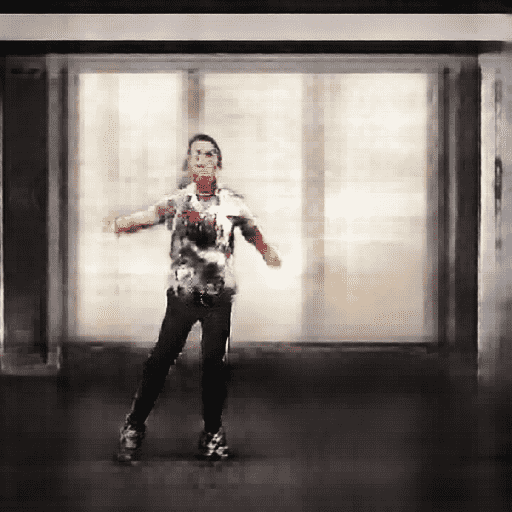} &
    \includegraphics[width=0.25\linewidth]{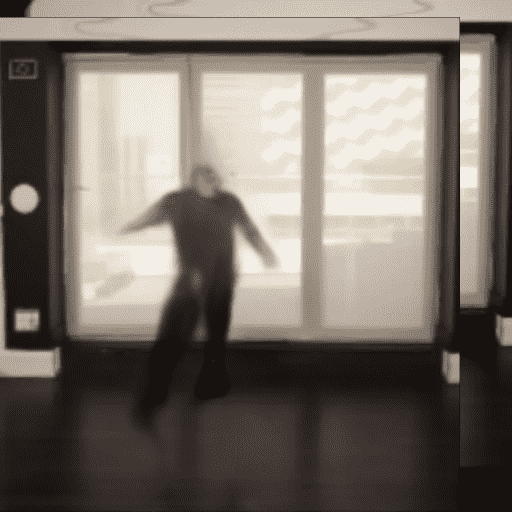} &
    \includegraphics[width=0.25\linewidth]{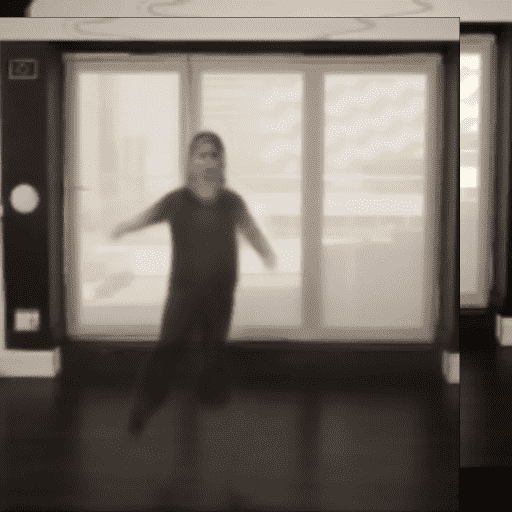} \\
    \end{tabular}
    }  
    \caption{
    {\bf Qualitative comparison.}
    Here we compare \method{} with baselines and network architectures.
    The Pix2PixHD $K=\infty$ model is an upper bound tuned on
    {\em all} available frames for 
    personalization. The last four columns compare the 
    constrained setting, where at test time only $K=5$
    frames are used for personalization, over 
    $T=20$ iterations of fine-tuning. We visualize results of challenging pose-image test 
    pairs that include rare poses not commonly scene in training.                                                                                                                                                                                                                                                                                                                                                                                                                                                                                                                                                                                                                                                                                                                                                                                                                                                                                                                                                                                                                                                                         
    In this case, the pre-trained baseline tends to copy clothing and backgrounds 
    from the training set, while \method{} is much better at personalizing to the 
    fine-tuning frames. The last two columns show that \posewarp{} is a more accurate 
    base architecture than Pix2PixHD, but meta-learning on top of it still produces more accurate 
    images with less blur.}

    \label{fig:improve}
\end{figure}

We start by building our baseline retargeting model, 
based on Pix2PixHD~\citep{wang2018pix2pixHD,chan2018everybody}.
To get a sense of the upper bound performance of our model, 
we train the model for each test video with no constraints 
on $T$ or $K$, starting from the model pre-trained on our train set. 
Specifically, we use all the frames from the first 85\%
of each video, and train it for 10 epochs.
We report the performance of this model in first section of Table~\ref{tab:baseline}
and show sample generations in second column of Figure~\ref{fig:improve}.
Since this model gets strong quantitative and qualitative performance,
we stick with it as our base retargeting architecture through the 
rest of the experiments. We also employ a baseline retargeting model based on \posewarp{} for evaluation,
but we focus on \poseim{} for further experimentation
due to its relative simplicity.

Now we evaluate the performance of our model in constrained settings, where we 
want to learn to personalize given a few samples and in a constrained 
computational budget. Hence, we use a pretrained model on train set and a random model,
and we personalize them by finetuning on each test video. 
As Table~\ref{tab:baseline} shows, applying constraints leads to a drop
in performance in all methods, as expected from using only 5 frames finetuned over 20 iterations.
Finally, we compare that to the \method{} model: in that case, we start from
the pre-trained model, and do meta-learning on top of those parameters to
optimize them for the transfer task as described in Section~\ref{sec:approach}.
That leads to a significant improvement over the pretrained model,
showing the strength of \method{} for this task.

In Figure~\ref{fig:improve}\footnote{Video visualization at \url{https://youtu.be/NlUmsd9aU-4}}, we visualize the predictions using the 
unconstrained model, as well as the constrained models trained 
using \method{} and without, i.e.\ with simple pretraining.
It is interesting to note that the 
meta-learned model is able to adapt to the color of the clothing and
the background much better than a pretrained model, given the 
same frames for personalization. This reinforces \method{} is a much 
better initialization for few-shot personalization than directly
finetuning from a generic pretrained model. We further explore this quality of
coherence in the next section. 

\vspace{-0.1in}
\subsection{Ablations}

We now ablate the key design choices in our \method{} formulation.
One of the strengths of our formulation is the explicit control on 
the supervision provided and computation the model is allowed to perform,
and depending on the use-case, those parameters can easily be tweaked.
We explore the effect of \method{} on those parameters next on the \poseim{} base retargeting architecture.

\begin{figure}[t]
    \centering
    \includegraphics[height = 1.5in]{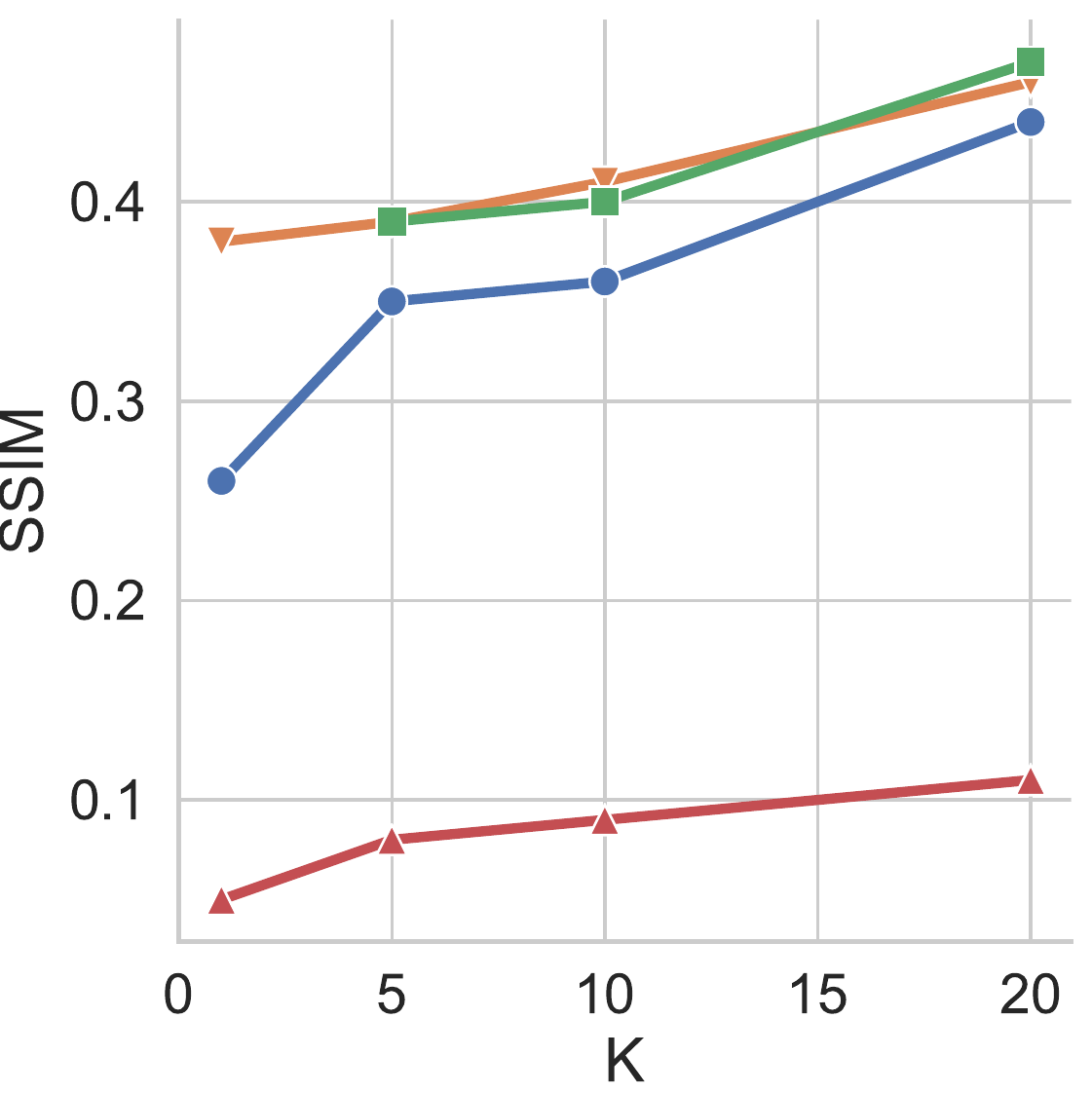}
    \includegraphics[height = 1.5in]{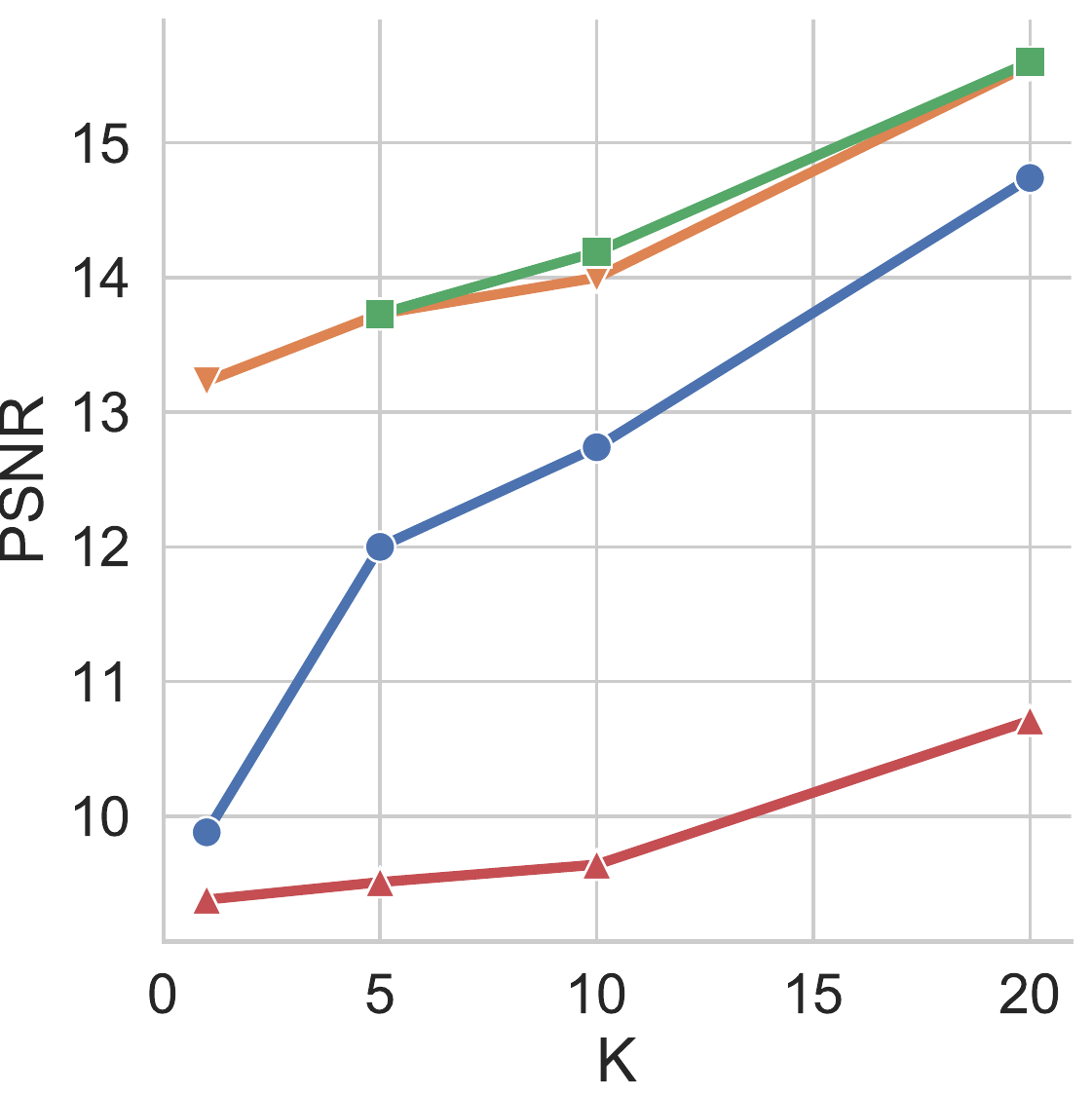}
    \includegraphics[height = 1.5in]{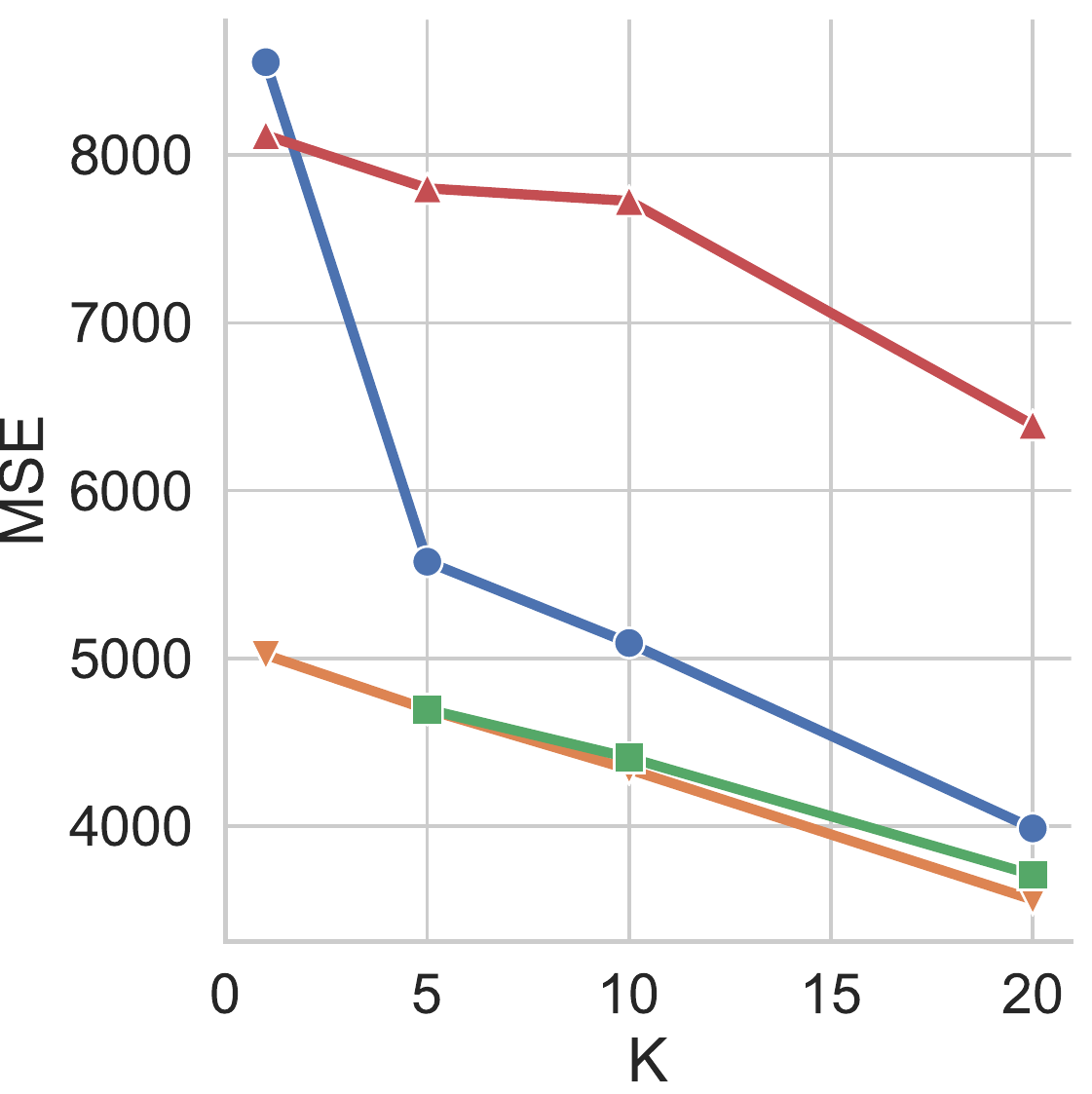}
    \includegraphics[height=1.5in,trim={10.3cm 1.1cm 0 0},clip]{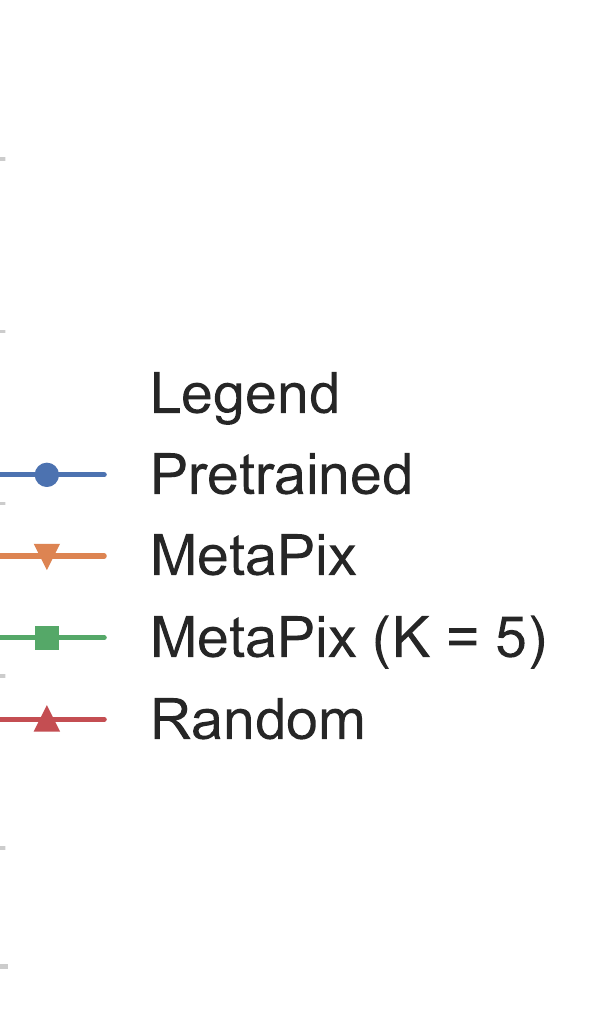}
    \caption{
    {\bf Personalization using $K$ frames.}
    We find that while all initializations get better with increasing $K$,
    and using \method{} consistently outperforms simple pretraining.
    Moreover we note that even using a model trained with \method{} for $K=5$
    works well at any $K$ value used at test time, showing the generalizability of \method{}.
    It is worth noting that the biggest gap is seen at lower values of $K$,
    showing our method is most useful in cases where one has little 
    data for personalization.
    }\label{fig:var_k}
\end{figure}

{\noindent \bf Variation in $K$:}
We vary the amount of supervision for personalization, $K$, and evaluate
its effect on the metrics in Figure~\ref{fig:var_k}. We compare the following models:
a) Randomly initialized, b) Pretrained on the train set, c) Trained using \method{}
for each value of $K$ and tested with the same $K$, and d) Trained using \method{} 
for $K=5$ and tested at each value of $K$. 
The last one tests the generalizability of \method{} to different values of $K$ 
at train and test time. We find that the \method{} trained models consistently
perform better than a simple pretrained model on all metrics. Notably, the model only trained 
for $K=5$ is still able to obtain strong performance at different $K$ values,
showing the \method{} trained model can generalize beyond the specific 
setup it is optimized for. The gap between the \method{} trained model and 
the pretrained model tends to reduce with higher $K$, which is as expected: 
more data for personalization would likely reduce the importance of the 
initialization. However, there is a clear and significant gap for lower 
values of $K$, showing that \method{} is highly effective for retargeting 
from few samples. In fact, we find that meta-learning is most effective for $K=1$, corresponding to the challenging scenario of video-to-image retargeting.

\begin{figure}[t]
    \centering
    \includegraphics[height = 1.5in]{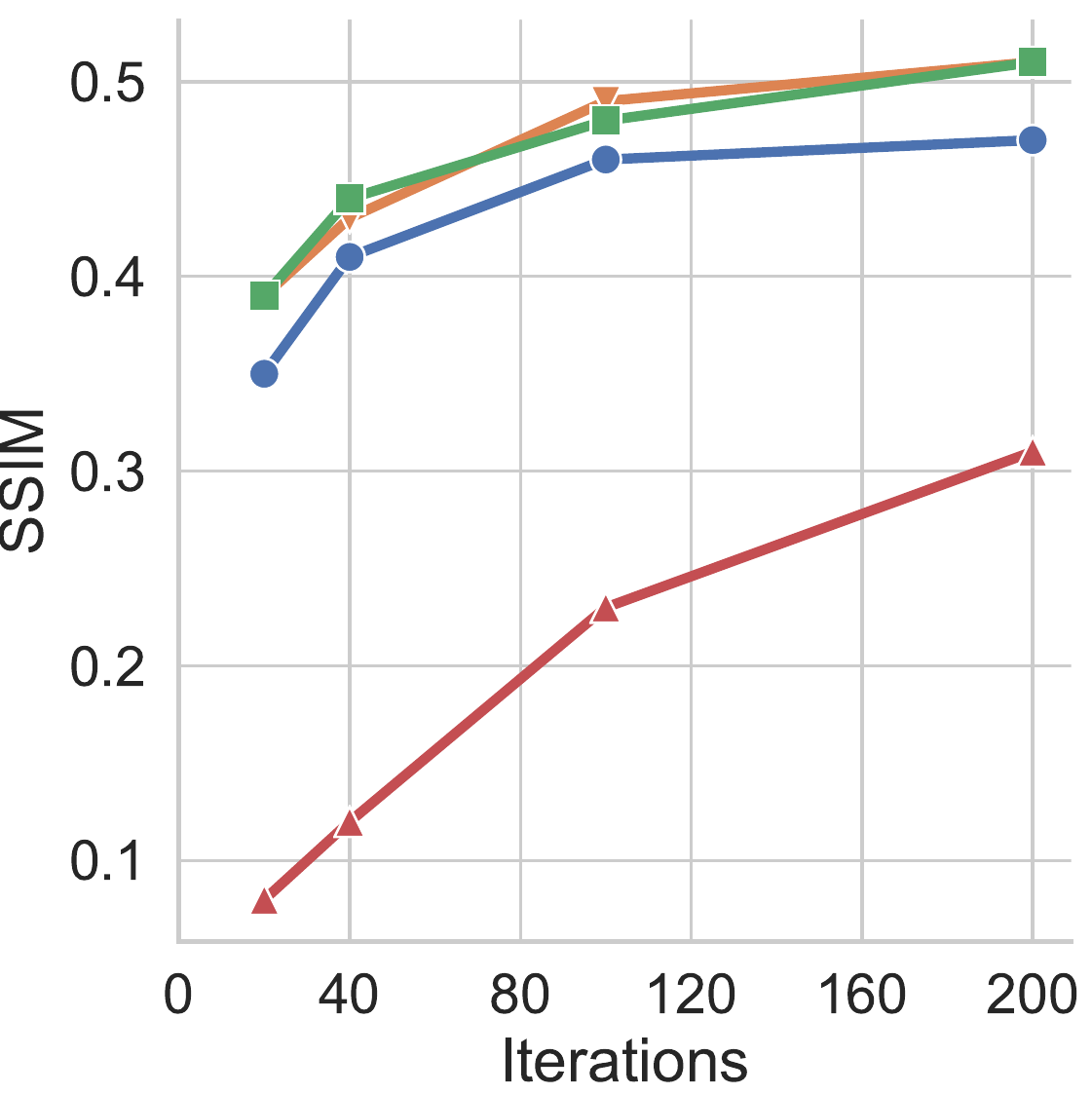}
    \includegraphics[height = 1.5in]{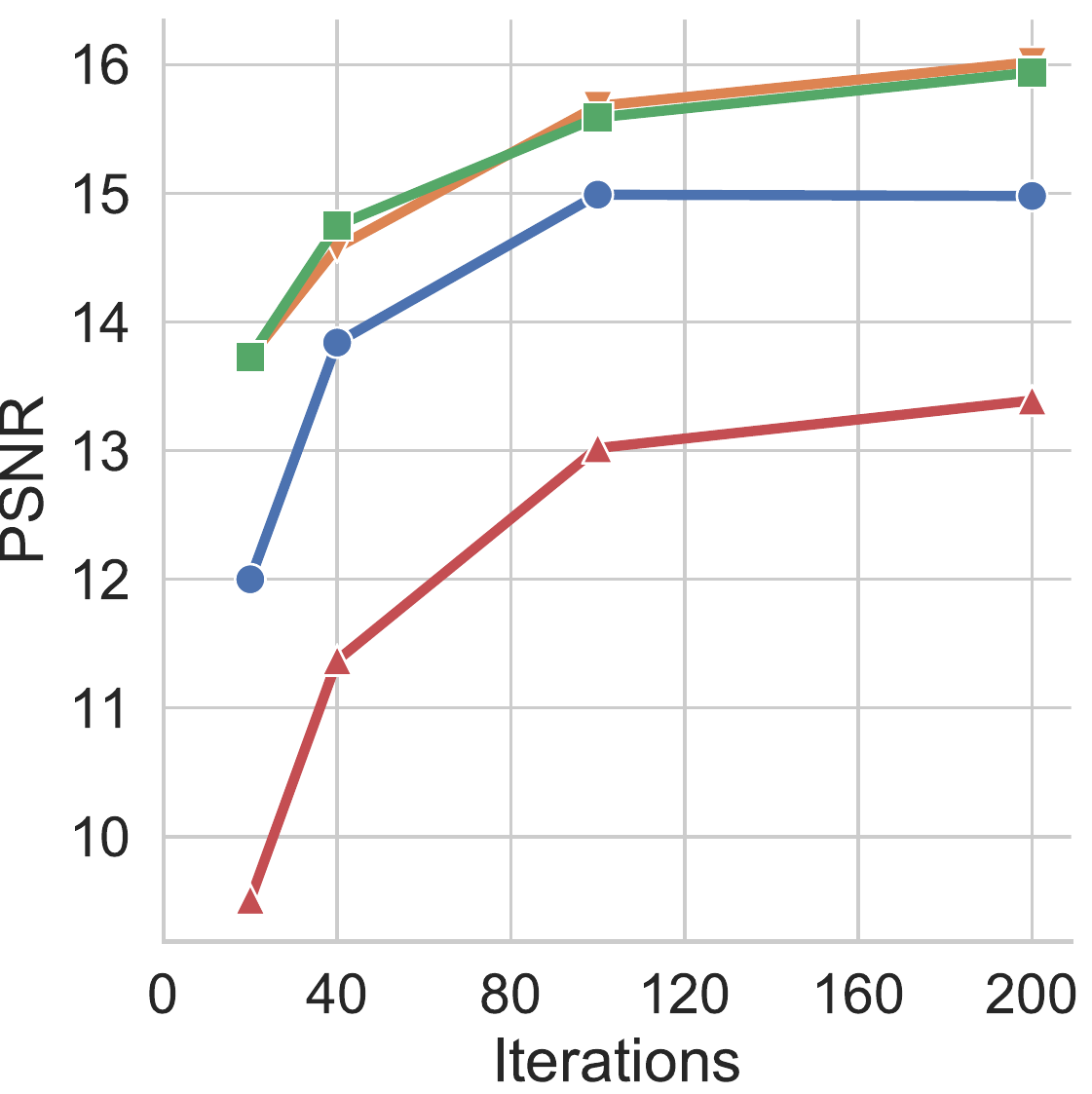}
    \includegraphics[height = 1.5in]{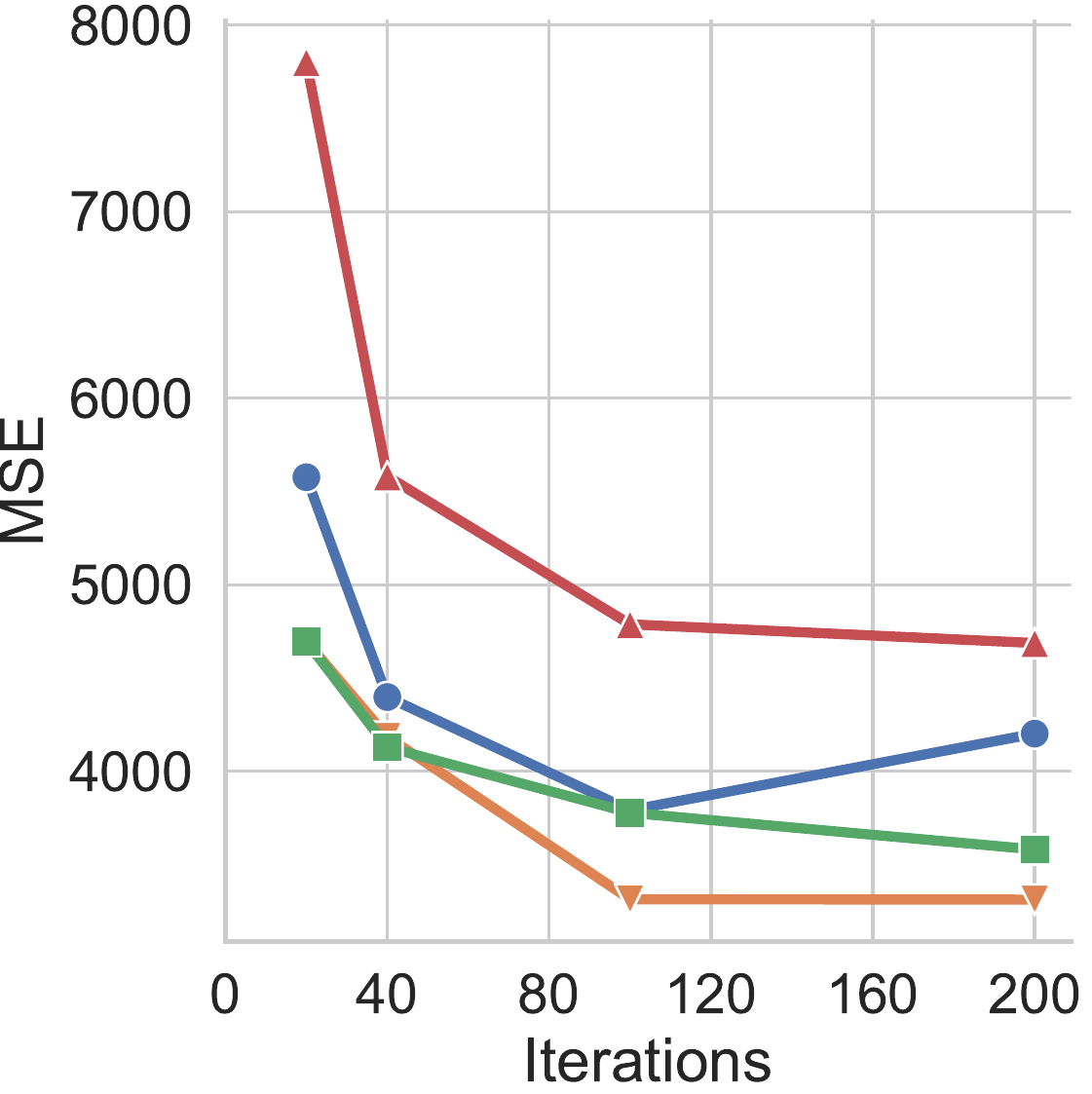}
    \includegraphics[height=1.5in,trim={10.3cm 1cm 0 0},clip]{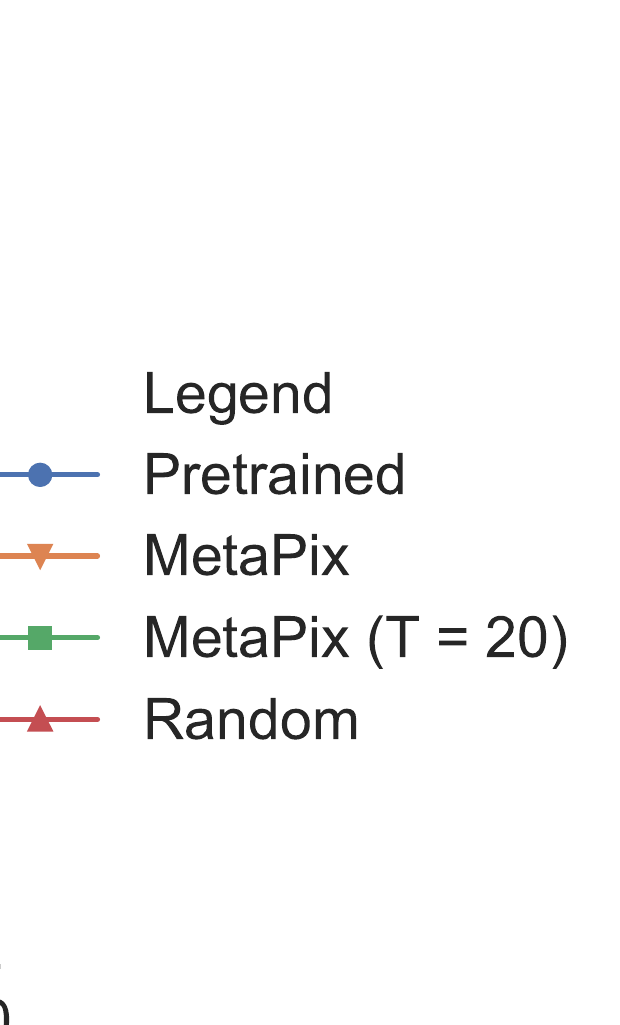}
    \caption{
    {\bf Personalization after $T$ iterations.}
    We compare performance on increasing the computational budget for personalization.
    As expected all initializations improve with $T$, though \method{} consistently outperforms
    random or pretraining. Again we see strong generalizability, as a \method{} model
    trained for $T=20$ performs well at other $T$ values used at test time.
    }\label{fig:var_t}
\end{figure}

{\noindent \bf Variation in $T$:}
Similar to variation in supervision, we experiment with varying the computation, or $T$,
in Figure~\ref{fig:var_t}. We experiment with a similar set of baselines as in the 
case for $K$, and again observe that the \method{} model consistently outperforms
random initialization or pretraining on all metrics. Also, we see similar generalizability,
as the model metatrained for $T=20$ is able to perform well for other $T$ values 
at test time too. The ability for MetaPix to generalize across $K$ and $T$ implies cost-effective strategies for training. The computational cost for training a meta-learner is dominated by fine-tuning, which scales linearly with $K$ and $T$. Training with smaller values of both can result in significant speedups -- up to 10$\times$ in our experiments. 

{\noindent \bf Variation of meta learning rate $\epsilon$:}
We also experimented with changing the meta learning rate. At $\epsilon=0.1$ ($K=5, T=200$), we obtained
SSIM=0.47, similar to what the pretrained model gets. Using our default $\epsilon=1.0$, improves performance 
to 0.51. Hence, a higher meta learning rate was imperative to see improvements with \method{}.

{\noindent \bf Only training the generator:}
We apply Reptile in a GAN setting, where we jointly 
meta-optimize two networks. We also experimented with freezing one of the networks,
specifically the discriminator, to the weights learned during pretraining. For our $K=5,T=200,\epsilon=1.0$ setup, we obtain similar performance as optimizing both,
suggesting that a `universal' discriminator might suffice for meta-learning on GANs.

\begin{figure}[t]
    \centering
    \includegraphics[width=\linewidth]{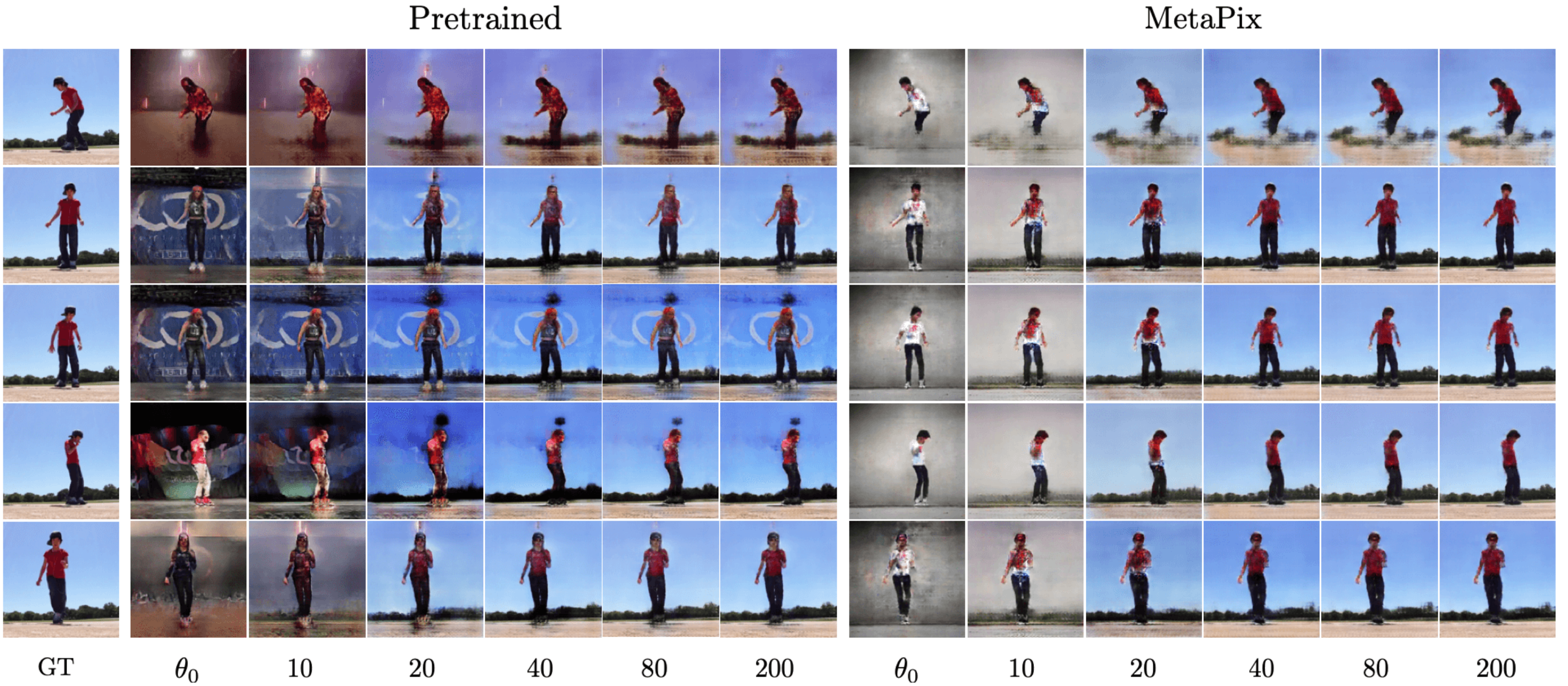}
    \caption{
        {\bf Visualizing finetuning between \method{} and Pretrained.} 
        We compare \method{}'s initialization for the $K=5, T=200$ task to our pretrained model initialization. 
        We visualize models obtained during iterations of finetuning, at $0,10,20,40,80$ and $200$ iterations for 5 random test pose-image pairs.
        The images generated by \method{}'s initialization are temporally coherent, 
        whereas the pretrained weights produce various training images depending on the pose.
        As observed in the intermediate iterations, the initialization translates 
        its temporal coherence properties across finetuning iterations as well. 
        This further reinforces our belief that \method{} 
        learns an initialization that is able to 
        quickly adapt to the actor and background appearance from the few samples provided at test time.
}\label{fig:temp_coherence_ft}
    \vspace{-0.1in}
\end{figure}

{\noindent \bf Visualizing the dynamics of personalization:}
In order to examine the process of personalization, we visualize models obtained during iterations of finetuning, at $10,20,40,80$ and $200$ iterations for 5 random test pose-image pairs.
We compare both the pretrained and metalearned model, trained for $k=5,T=200$.
Figure ~\ref{fig:temp_coherence_ft}\footnote{Video visualization at \url{https://youtu.be/bxJJXCK4IoQ}} shows images generated by these intermediate iterations. 
Both methods learn clothing details and background colors after $20$ iterations. 
Interestingly, \method{} produces images that are temporally coherent, even upon initialization, while the pretrained baseline produces images whose
background and clothing vary with pose. This more coherent initialization appears to translate to more coherent generated images after personalization.

  \vspace{-0.1in}
\section{Conclusion} 
\vspace{-0.1in}
We have explored the task of quickly and efficiently retargeting human actions 
from one video to another, given a limited number of samples 
from the target domain. 
We formalize this as a few-shot personalization problem, where we
first learn a generic 
generative model on large 
amounts of data, and then specialize it to a small amount of target frames 
via finetuning. We further propose a
novel meta-learning based approach, \method{}, to learn this generic 
model in a way that is more amenable to personalization via fine-tuning. 
To do so, we repurpose a first-order meta-learning algorithm,
Reptile, to adversarially meta-optimize both the generator and discriminator of a generative adversarial network. 
We experiment with
it on in-the-wild YouTube videos, and find that \method{} outperforms widely-used approaches for pretraining,
while generating temporally coherent videos.

 {\bf Acknowledgements:} This research is based upon work supported in part by NSF Grant 1618903, the Intel Science and Technology Center for Visual Cloud Systems (ISTC-VCS), and Google. 
\bibliography{iclr2020_conference}

\begin{thebibliography}{42}
\providecommand{\natexlab}[1]{#1}
\providecommand{\url}[1]{\texttt{#1}}
\expandafter\ifx\csname urlstyle\endcsname\relax
  \providecommand{\doi}[1]{doi: #1}\else
  \providecommand{\doi}{doi: \begingroup \urlstyle{rm}\Url}\fi

\bibitem[Andrychowicz et~al.(2016)Andrychowicz, Denil, Gomez, Hoffman, Pfau,
  Schaul, Shillingford, and De~Freitas]{andrychowicz2016learning}
Marcin Andrychowicz, Misha Denil, Sergio Gomez, Matthew~W Hoffman, David Pfau,
  Tom Schaul, Brendan Shillingford, and Nando De~Freitas.
\newblock Learning to learn by gradient descent by gradient descent.
\newblock In \emph{NeurIPS}, 2016.

\bibitem[Balakrishnan et~al.(2018)Balakrishnan, Zhao, Dalca, Durand, and
  Guttag]{balakrishnan2018synthesizing}
Guha Balakrishnan, Amy Zhao, Adrian~V Dalca, Fredo Durand, and John Guttag.
\newblock Synthesizing images of humans in unseen poses.
\newblock In \emph{CVPR}, 2018.

\bibitem[Bansal et~al.(2018)Bansal, Ma, Ramanan, and Sheikh]{Recycle-GAN}
Aayush Bansal, Shugao Ma, Deva Ramanan, and Yaser Sheikh.
\newblock Recycle-gan: Unsupervised video retargeting.
\newblock In \emph{ECCV}, 2018.

\bibitem[Bart \& Ullman(2005)Bart and Ullman]{bart2005cross}
Evgeniy Bart and Shimon Ullman.
\newblock Cross-generalization: Learning novel classes from a single example by
  feature replacement.
\newblock In \emph{CVPR}, 2005.

\bibitem[Bertinetto et~al.(2016)Bertinetto, Henriques, Valmadre, Torr, and
  Vedaldi]{bertinetto2016learning}
Luca Bertinetto, Jo{\~a}o~F Henriques, Jack Valmadre, Philip Torr, and Andrea
  Vedaldi.
\newblock Learning feed-forward one-shot learners.
\newblock In \emph{NeurIPS}, 2016.

\bibitem[Brock et~al.(2019)Brock, Donahue, and Simonyan]{brock2018large}
Andrew Brock, Jeff Donahue, and Karen Simonyan.
\newblock Large scale gan training for high fidelity natural image synthesis.
\newblock In \emph{ICLR}, 2019.

\bibitem[Chan et~al.(2018)Chan, Ginosar, Zhou, and Efros]{chan2018everybody}
Caroline Chan, Shiry Ginosar, Tinghui Zhou, and Alexei~A Efros.
\newblock Everybody dance now.
\newblock \emph{arXiv preprint arXiv:1808.07371}, 2018.

\bibitem[Duan et~al.(2016)Duan, Schulman, Chen, Bartlett, Sutskever, and
  Abbeel]{duan2016rl}
Yan Duan, John Schulman, Xi~Chen, Peter~L Bartlett, Ilya Sutskever, and Pieter
  Abbeel.
\newblock Rl2: Fast reinforcement learning via slow reinforcement learning.
\newblock \emph{arXiv preprint arXiv:1611.02779}, 2016.

\bibitem[Fei-Fei et~al.(2006)Fei-Fei, Fergus, and Perona]{fei2006one}
Li~Fei-Fei, Rob Fergus, and Pietro Perona.
\newblock One-shot learning of object categories.
\newblock \emph{TPAMI}, 2006.

\bibitem[Finn et~al.(2016)Finn, Goodfellow, and Levine]{finn2016unsupervised}
Chelsea Finn, Ian Goodfellow, and Sergey Levine.
\newblock Unsupervised learning for physical interaction through video
  prediction.
\newblock In \emph{NeurIPS}, 2016.

\bibitem[Finn et~al.(2017)Finn, Abbeel, and Levine]{finn2017model}
Chelsea Finn, Pieter Abbeel, and Sergey Levine.
\newblock Model-agnostic meta-learning for fast adaptation of deep networks.
\newblock In \emph{ICML}, 2017.

\bibitem[Gleicher(1998)]{gleicher1998retargetting}
Michael Gleicher.
\newblock Retargetting motion to new characters.
\newblock In \emph{SIGGRAPH}, 1998.

\bibitem[Goodfellow et~al.(2014)Goodfellow, Pouget-Abadie, Mirza, Xu,
  Warde-Farley, Ozair, Courville, and Bengio]{goodfellow2014generative}
Ian Goodfellow, Jean Pouget-Abadie, Mehdi Mirza, Bing Xu, David Warde-Farley,
  Sherjil Ozair, Aaron Courville, and Yoshua Bengio.
\newblock Generative adversarial nets.
\newblock In \emph{NeurIPS}, 2014.

\bibitem[Hariharan \& Girshick(2017)Hariharan and Girshick]{hariharan2017low}
Bharath Hariharan and Ross Girshick.
\newblock Low-shot visual recognition by shrinking and hallucinating features.
\newblock In \emph{ICCV}, 2017.

\bibitem[Hochreiter et~al.(2001)Hochreiter, Younger, and
  Conwell]{hochreiter2001learning}
Sepp Hochreiter, A~Steven Younger, and Peter~R Conwell.
\newblock Learning to learn using gradient descent.
\newblock In \emph{ICANN}, 2001.

\bibitem[Hoffman et~al.(2018)Hoffman, Tzeng, Park, Zhu, Isola, Saenko, Efros,
  and Darrell]{Hoffman_cycada2017}
Judy Hoffman, Eric Tzeng, Taesung Park, Jun-Yan Zhu, Phillip Isola, Kate
  Saenko, Alexei~A. Efros, and Trevor Darrell.
\newblock Cycada: Cycle consistent adversarial domain adaptation.
\newblock In \emph{ICML}, 2018.

\bibitem[Isola et~al.(2017)Isola, Zhu, Zhou, and Efros]{pix2pix2016}
Phillip Isola, Jun-Yan Zhu, Tinghui Zhou, and Alexei~A Efros.
\newblock Image-to-image translation with conditional adversarial networks.
\newblock In \emph{CVPR}, 2017.

\bibitem[Johnson et~al.(2016)Johnson, Alahi, and
  Fei-Fei]{johnson2016perceptual}
Justin Johnson, Alexandre Alahi, and Li~Fei-Fei.
\newblock Perceptual losses for real-time style transfer and super-resolution.
\newblock In \emph{ECCV}, 2016.

\bibitem[Kingma \& Ba(2015)Kingma and Ba]{kingma2015adam}
Diederik~P. Kingma and Jimmy~Lei Ba.
\newblock Adam: A method for stochastic optimization.
\newblock In \emph{ICLR}, 2015.

\bibitem[Kingma \& Welling(2014)Kingma and Welling]{kingma2013auto}
Diederik~P Kingma and Max Welling.
\newblock Auto-encoding variational bayes.
\newblock In \emph{ICLR}, 2014.

\bibitem[Liu et~al.(2018)Liu, Xu, Zollhoefer, Kim, Bernard, Habermann, Wang,
  and Theobalt]{liu2018neural}
Lingjie Liu, Weipeng Xu, Michael Zollhoefer, Hyeongwoo Kim, Florian Bernard,
  Marc Habermann, Wenping Wang, and Christian Theobalt.
\newblock Neural animation and reenactment of human actor videos.
\newblock \emph{arXiv preprint arXiv:1809.03658}, 2018.

\bibitem[Ma et~al.(2017)Ma, Jia, Sun, Schiele, Tuytelaars, and
  Van~Gool]{ma2017pose}
Liqian Ma, Xu~Jia, Qianru Sun, Bernt Schiele, Tinne Tuytelaars, and Luc
  Van~Gool.
\newblock Pose guided person image generation.
\newblock In \emph{NeurIPS}, 2017.

\bibitem[Misra et~al.(2017)Misra, Gupta, and Hebert]{misra2017red}
Ishan Misra, Abhinav Gupta, and Martial Hebert.
\newblock From red wine to red tomato: Composition with context.
\newblock In \emph{CVPR}, 2017.

\bibitem[Neverova et~al.(2018)Neverova, Alp~Guler, and
  Kokkinos]{neverova2018dense}
Natalia Neverova, Riza Alp~Guler, and Iasonas Kokkinos.
\newblock Dense pose transfer.
\newblock In \emph{ECCV}, 2018.

\bibitem[Nichol et~al.(2018)Nichol, Achiam, and Schulman]{nichol2018first}
Alex Nichol, Joshua Achiam, and John Schulman.
\newblock On first-order meta-learning algorithms.
\newblock \emph{arXiv preprint arXiv:1803.02999}, 2018.

\bibitem[Ronneberger et~al.(2015)Ronneberger, Fischer, and Brox]{unet}
Olaf Ronneberger, Philipp Fischer, and Thomas Brox.
\newblock U-net: Convolutional networks for biomedical image segmentation.
\newblock \emph{CoRR}, abs/1505.04597, 2015.
\newblock URL \url{http://arxiv.org/abs/1505.04597}.

\bibitem[Salakhutdinov et~al.(2012)Salakhutdinov, Tenenbaum, and
  Torralba]{salakhutdinov2012one}
Ruslan Salakhutdinov, Joshua Tenenbaum, and Antonio Torralba.
\newblock One-shot learning with a hierarchical nonparametric bayesian model.
\newblock In \emph{ICML Workshop}, 2012.

\bibitem[Santoro et~al.(2016)Santoro, Bartunov, Botvinick, Wierstra, and
  Lillicrap]{santoro2016meta}
Adam Santoro, Sergey Bartunov, Matthew Botvinick, Daan Wierstra, and Timothy
  Lillicrap.
\newblock Meta-learning with memory-augmented neural networks.
\newblock In \emph{ICML}, 2016.

\bibitem[Siarohin et~al.(2018)Siarohin, Sangineto, Lathuili{\`e}re, and
  Sebe]{siarohin2018deformable}
Aliaksandr Siarohin, Enver Sangineto, St{\'e}phane Lathuili{\`e}re, and Nicu
  Sebe.
\newblock Deformable gans for pose-based human image generation.
\newblock In \emph{CVPR}, 2018.

\bibitem[Simonyan \& Zisserman(2015)Simonyan and Zisserman]{Simonyan_14a}
Karen Simonyan and Andrew Zisserman.
\newblock Very deep convolutional networks for large-scale image recognition.
\newblock In \emph{ICLR}, 2015.

\bibitem[Thrun(1996)]{thrun1996learning}
Sebastian Thrun.
\newblock Is learning the n-th thing any easier than learning the first?
\newblock In \emph{NeurIPS}, 1996.

\bibitem[Vondrick et~al.(2016)Vondrick, Pirsiavash, and
  Torralba]{vondrick2016generating}
Carl Vondrick, Hamed Pirsiavash, and Antonio Torralba.
\newblock Generating videos with scene dynamics.
\newblock In \emph{NeurIPS}, 2016.

\bibitem[Walker et~al.(2017)Walker, Marino, Gupta, and Hebert]{walker2017pose}
Jacob Walker, Kenneth Marino, Abhinav Gupta, and Martial Hebert.
\newblock The pose knows: Video forecasting by generating pose futures.
\newblock In \emph{ICCV}, 2017.

\bibitem[Wang et~al.(2018{\natexlab{a}})Wang, Liu, Zhu, Liu, Tao, Kautz, and
  Catanzaro]{wang2018vid2vid}
Ting-Chun Wang, Ming-Yu Liu, Jun-Yan Zhu, Guilin Liu, Andrew Tao, Jan Kautz,
  and Bryan Catanzaro.
\newblock Video-to-video synthesis.
\newblock In \emph{NeurIPS}, 2018{\natexlab{a}}.

\bibitem[Wang et~al.(2018{\natexlab{b}})Wang, Liu, Zhu, Tao, Kautz, and
  Catanzaro]{wang2018pix2pixHD}
Ting-Chun Wang, Ming-Yu Liu, Jun-Yan Zhu, Andrew Tao, Jan Kautz, and Bryan
  Catanzaro.
\newblock High-resolution image synthesis and semantic manipulation with
  conditional gans.
\newblock In \emph{CVPR}, 2018{\natexlab{b}}.

\bibitem[Wang \& Hebert(2016{\natexlab{a}})Wang and Hebert]{wang2016learning}
Yu-Xiong Wang and Martial Hebert.
\newblock Learning from small sample sets by combining unsupervised
  meta-training with cnns.
\newblock In \emph{NeurIPS}, 2016{\natexlab{a}}.

\bibitem[Wang \& Hebert(2016{\natexlab{b}})Wang and
  Hebert]{wang2016learning_to_learn}
Yu-Xiong Wang and Martial Hebert.
\newblock Learning to learn: Model regression networks for easy small sample
  learning.
\newblock In \emph{ECCV}, 2016{\natexlab{b}}.

\bibitem[Wang et~al.(2017)Wang, Ramanan, and
  Hebert]{wang2017learning_model_tail}
Yu-Xiong Wang, Deva Ramanan, and Martial Hebert.
\newblock Learning to model the tail.
\newblock In \emph{NeurIPS}, 2017.

\bibitem[Wang et~al.(2018{\natexlab{c}})Wang, Girshick, Hebert, and
  Hariharan]{wang2018low}
Yu-Xiong Wang, Ross Girshick, Martial Hebert, and Bharath Hariharan.
\newblock Low-shot learning from imaginary data.
\newblock In \emph{CVPR}, 2018{\natexlab{c}}.

\bibitem[Zakharov et~al.(2019)Zakharov, Shysheya, Burkov, and
  Lempitsky]{zakharov2019few}
Egor Zakharov, Aliaksandra Shysheya, Egor Burkov, and Victor Lempitsky.
\newblock Few-shot adversarial learning of realistic neural talking head
  models.
\newblock \emph{arXiv preprint arXiv:1905.08233}, 2019.

\bibitem[Zhou et~al.(2019)Zhou, Wang, Fang, Bui, and Berg]{zhou2019dance}
Yipin Zhou, Zhaowen Wang, Chen Fang, Trung Bui, and Tamara~L. Berg.
\newblock Dance dance generation: Motion transfer for internet videos.
\newblock \emph{arXiv preprint arXiv:1904.00129}, 2019.

\bibitem[Zhu et~al.(2017)Zhu, Park, Isola, and Efros]{CycleGAN2017}
Jun-Yan Zhu, Taesung Park, Phillip Isola, and Alexei~A Efros.
\newblock Unpaired image-to-image translation using cycle-consistent
  adversarial networks.
\newblock In \emph{ICCV}, 2017.

\end{thebibliography}
\bibliographystyle{iclr2020_conference}

\appendix
\section{Appendix}

\begin{figure}[t]
  \centering
  \includegraphics[width=\linewidth]{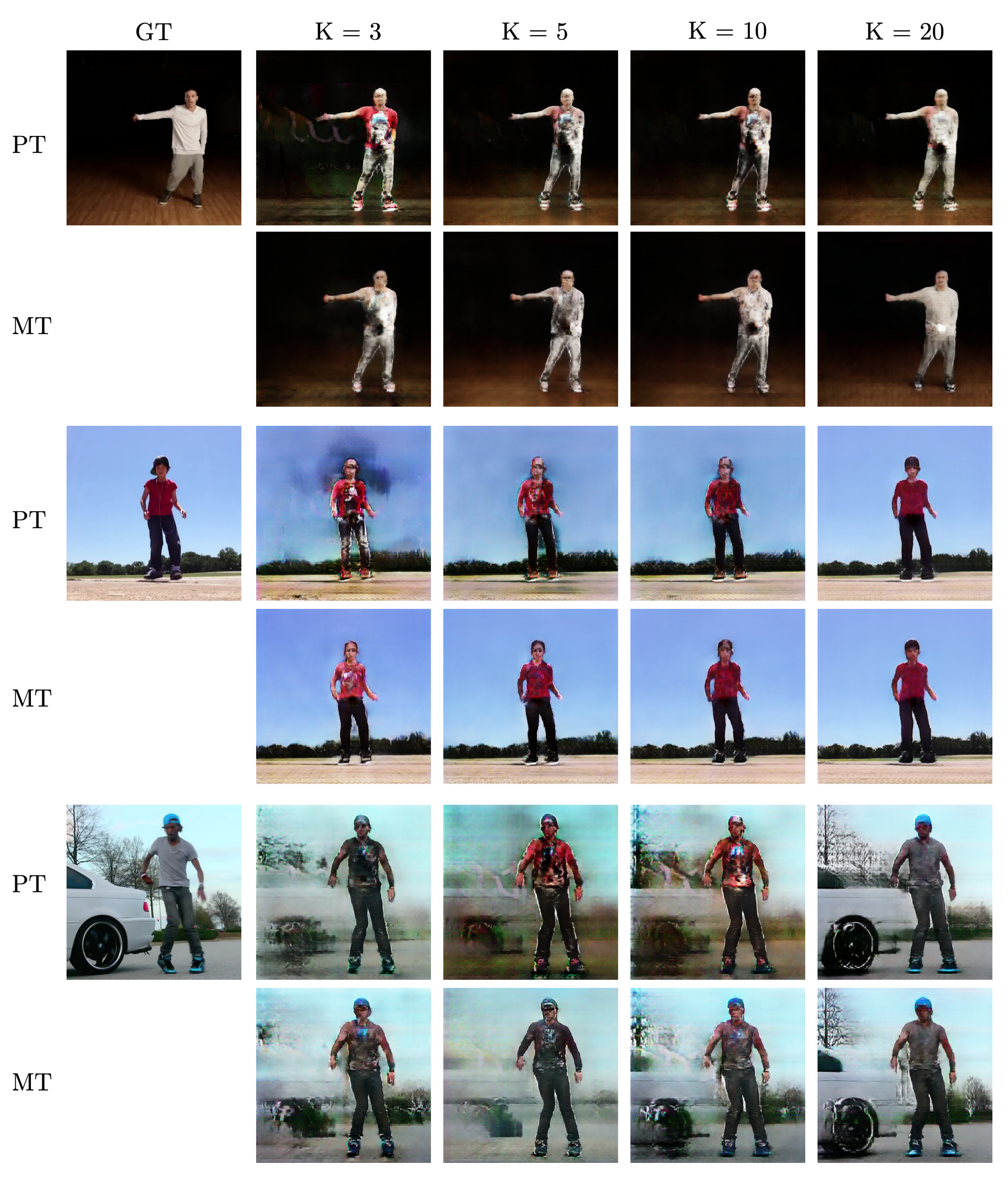}
  \caption{
      {\bf Qualitative Variation in K.}
      We compare the MetaPix-trained models (MT) with their pretrained counterparts (PT) for $K = [3, 5, 10, 20]$.  
      We fix the base architecture to Pix2PixHD and $T = 20$. With higher $K$, both methods generate good images, 
      but with lower $K$, \method{} generates backgrounds and clothing that better match the ground-truth. 
      Our results illustrate that meta-learning excels in the few-shot regime.}
    
      \label{fig:k_fig}
\end{figure}

\begin{figure}
  \centering
  \setlength{\tabcolsep}{2pt}
  \resizebox{\linewidth}{!}{
  \begin{tabular}{cccccc}
  Ground Truth & PT & MT T = 20 &  MT T = 40 & MT T = 100 & MT T = 200 \\
  \includegraphics[width=0.25\linewidth]{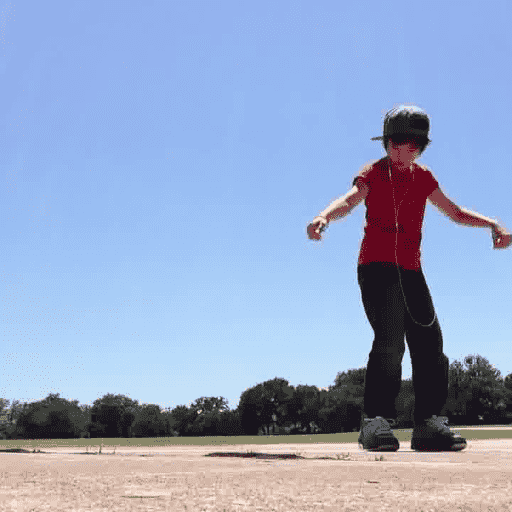} &
  \includegraphics[width=0.25\linewidth]{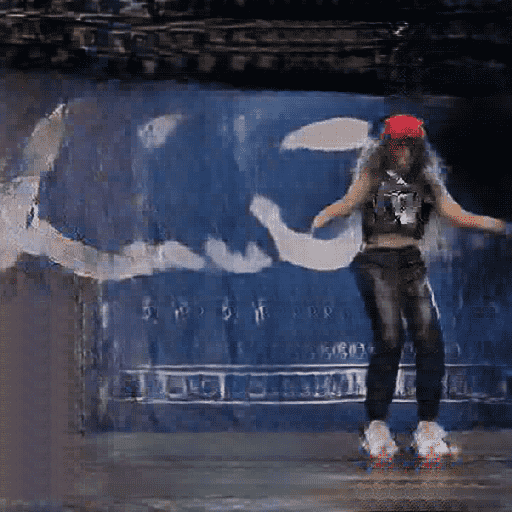} & 
  \includegraphics[width=0.25\linewidth]{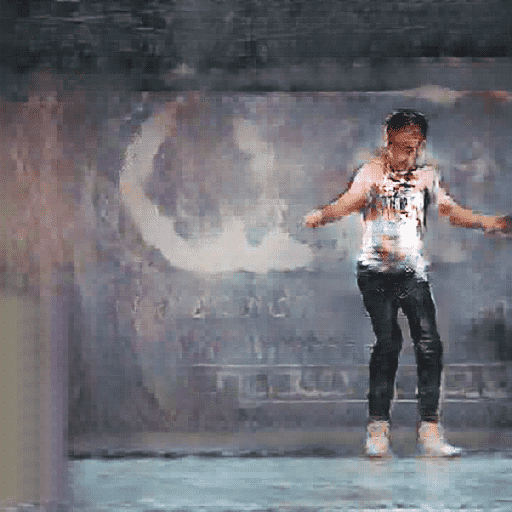} & 
  \includegraphics[width=0.25\linewidth]{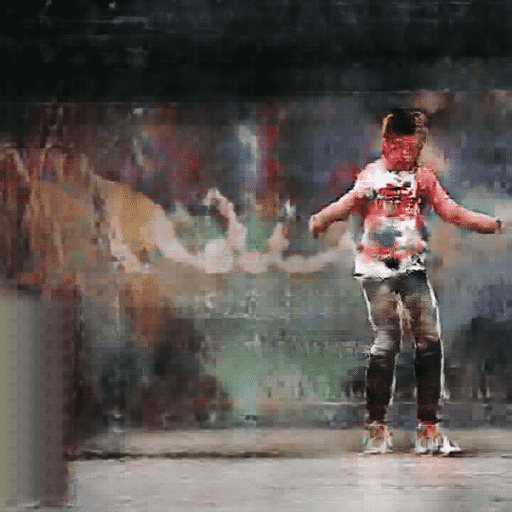} &
  \includegraphics[width=0.25\linewidth]{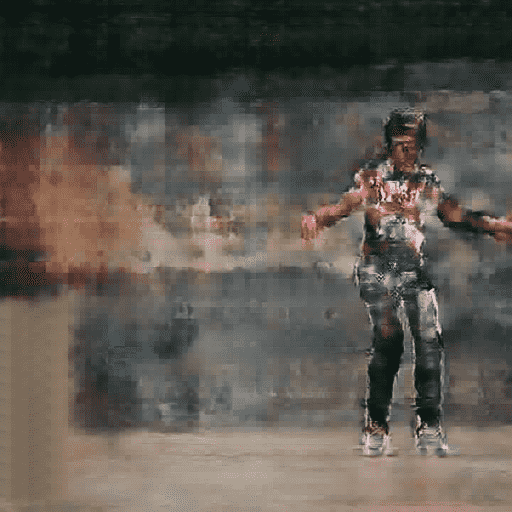} &
  \includegraphics[width=0.25\linewidth]{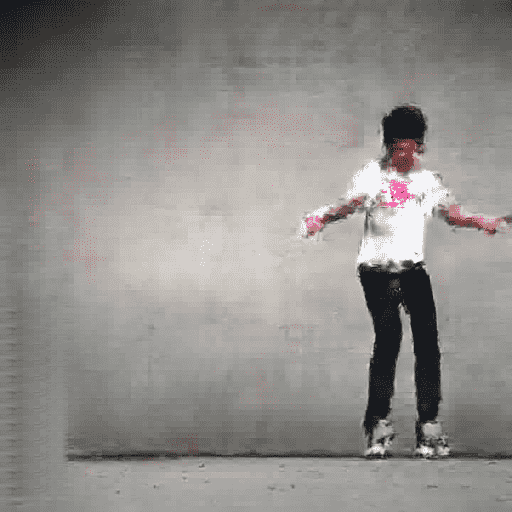} \\
  \includegraphics[width=0.25\linewidth]{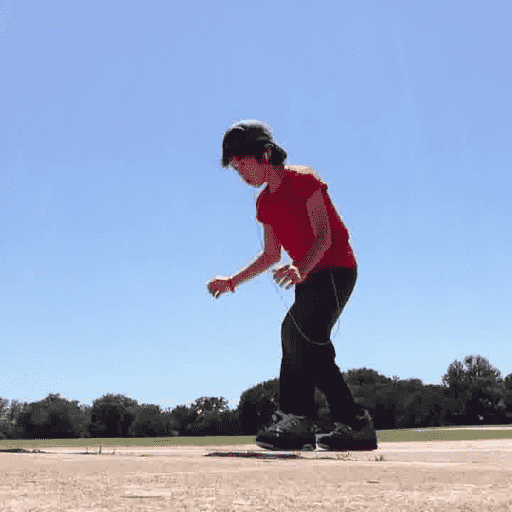} &
  \includegraphics[width=0.25\linewidth]{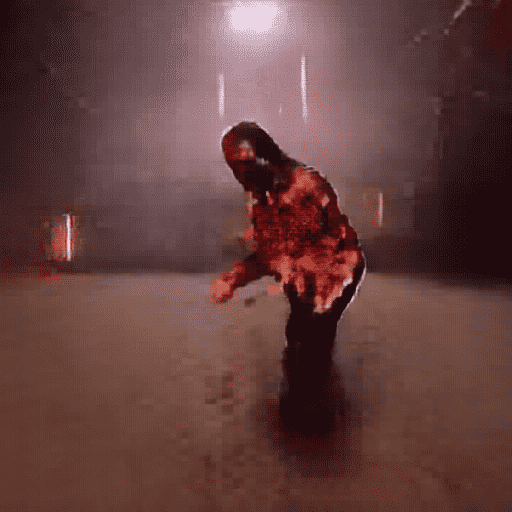} & 
  \includegraphics[width=0.25\linewidth]{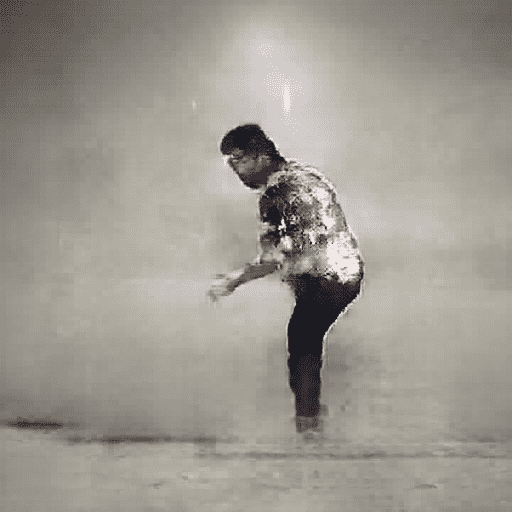} & 
  \includegraphics[width=0.25\linewidth]{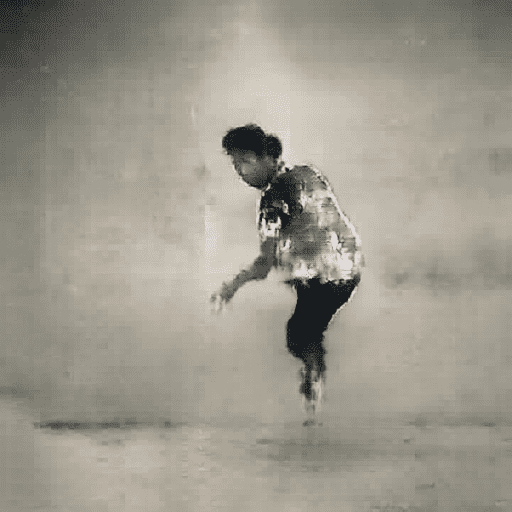} &
  \includegraphics[width=0.25\linewidth]{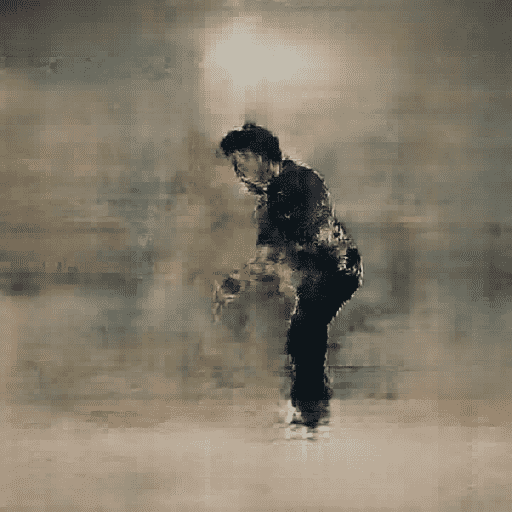} &
  \includegraphics[width=0.25\linewidth]{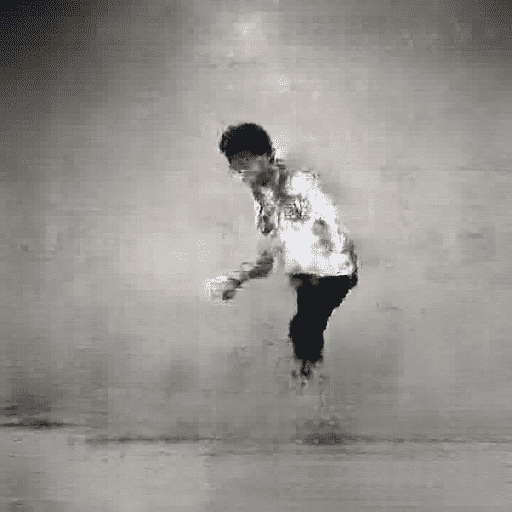} \\
  \includegraphics[width=0.25\linewidth]{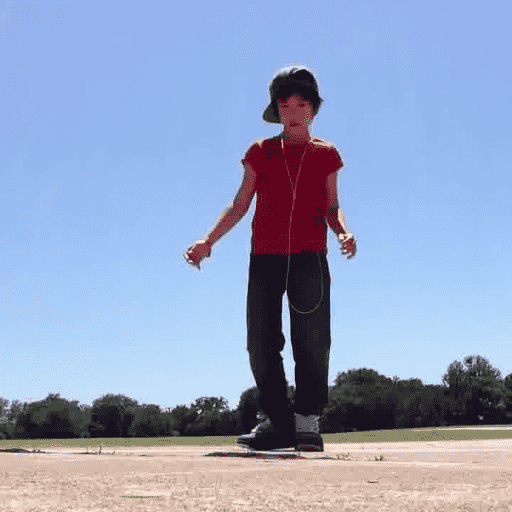} &
  \includegraphics[width=0.25\linewidth]{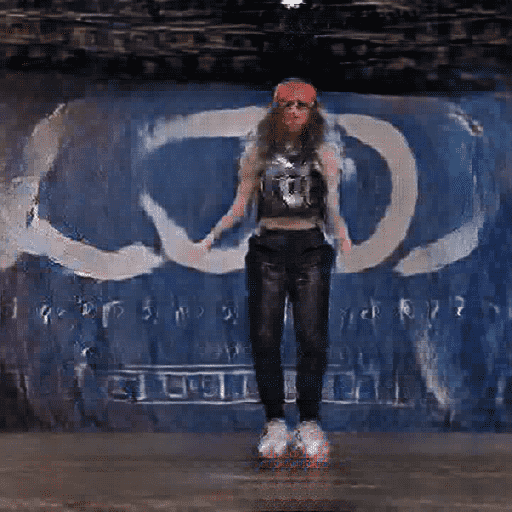} & 
  \includegraphics[width=0.25\linewidth]{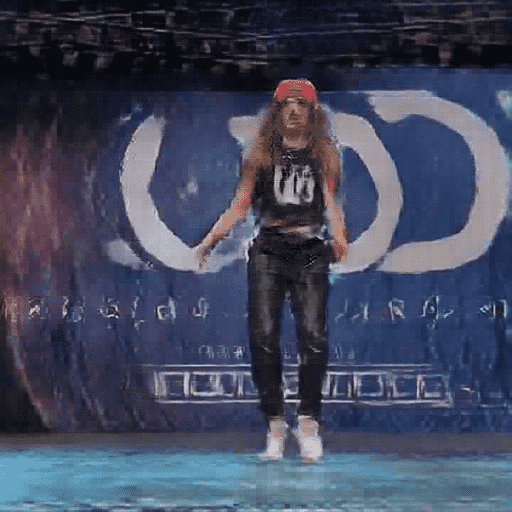} & 
  \includegraphics[width=0.25\linewidth]{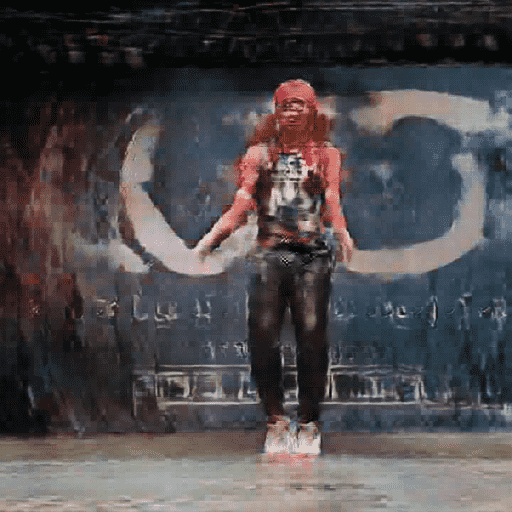} &
  \includegraphics[width=0.25\linewidth]{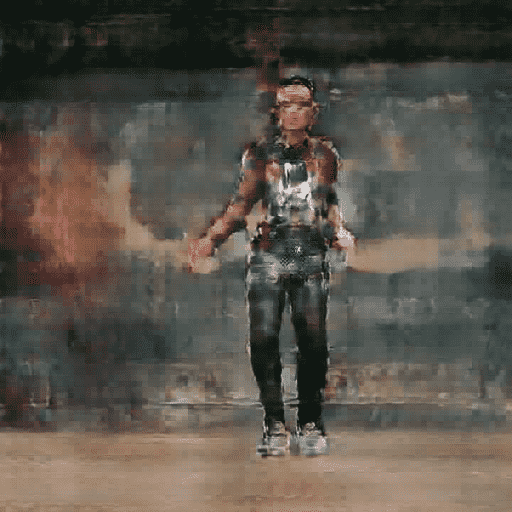} &
  \includegraphics[width=0.25\linewidth]{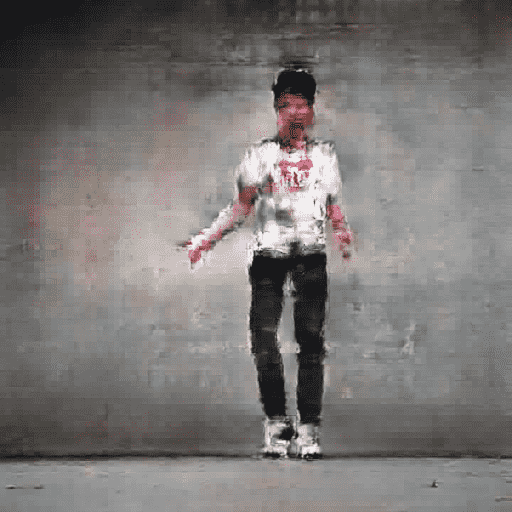} \\
  \includegraphics[width=0.25\linewidth]{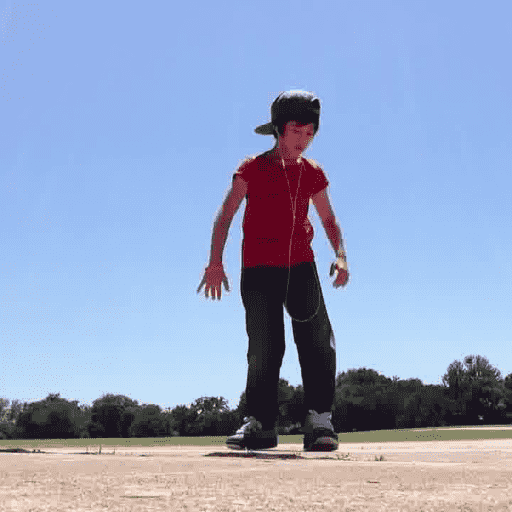} &
  \includegraphics[width=0.25\linewidth]{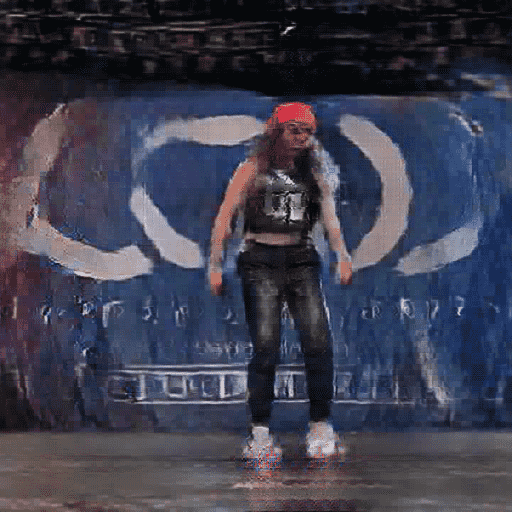} & 
  \includegraphics[width=0.25\linewidth]{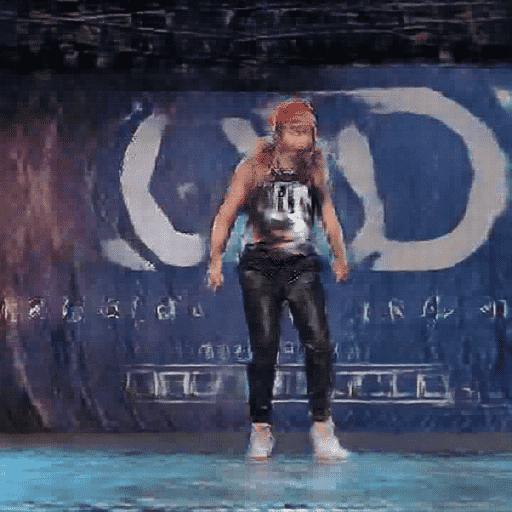} & 
  \includegraphics[width=0.25\linewidth]{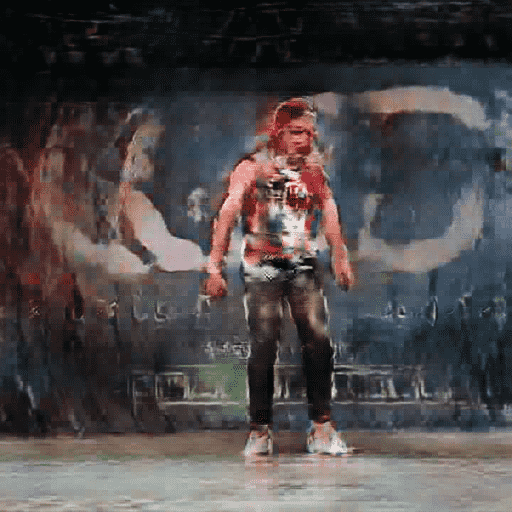} &
  \includegraphics[width=0.25\linewidth]{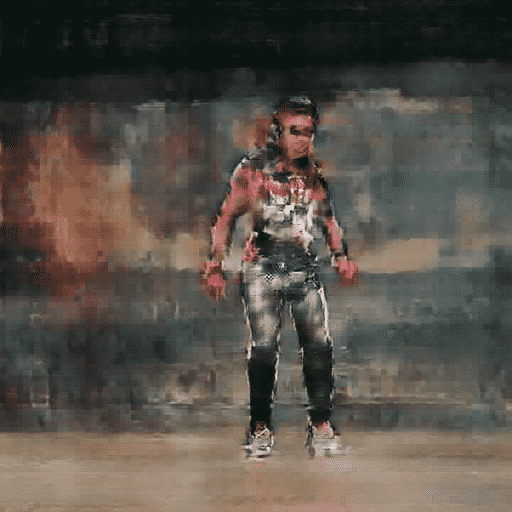} &
  \includegraphics[width=0.25\linewidth]{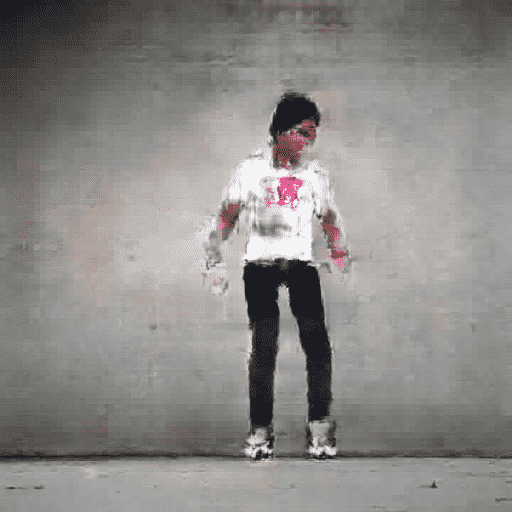} \\

  \includegraphics[width=0.25\linewidth]{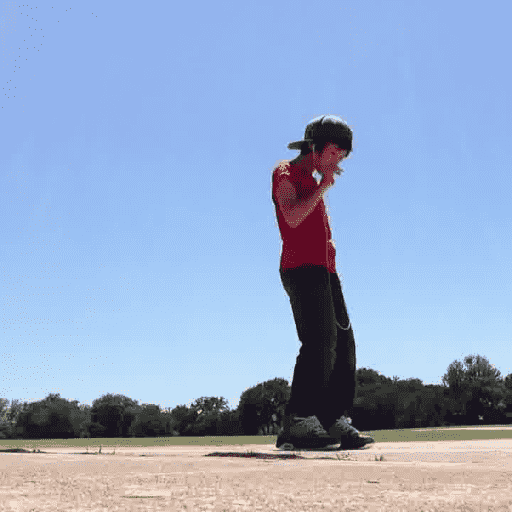} &
  \includegraphics[width=0.25\linewidth]{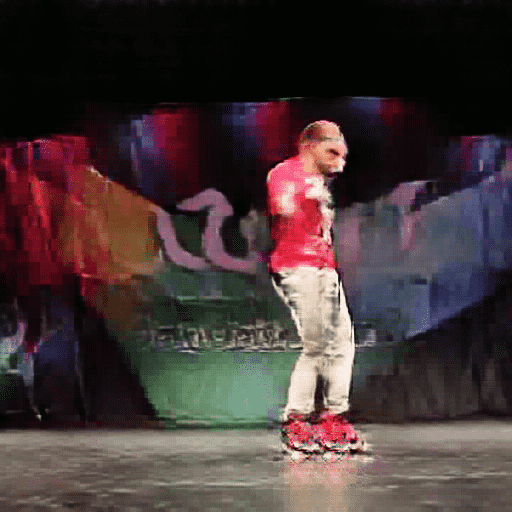} & 
  \includegraphics[width=0.25\linewidth]{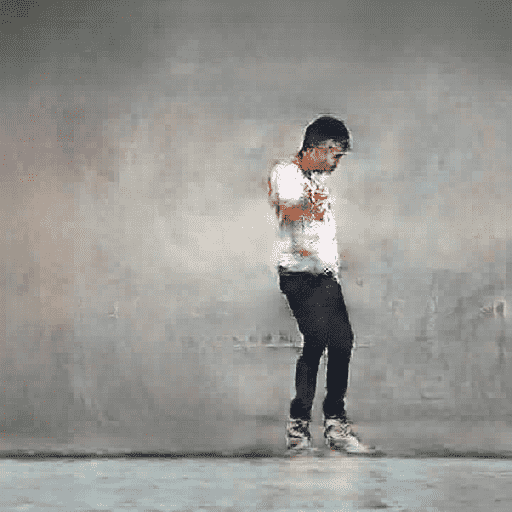} & 
  \includegraphics[width=0.25\linewidth]{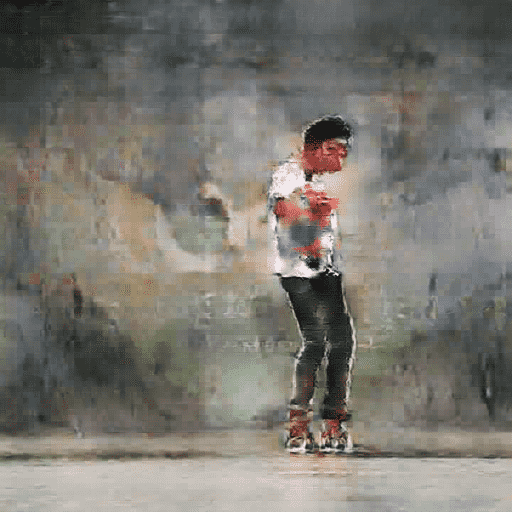} &
  \includegraphics[width=0.25\linewidth]{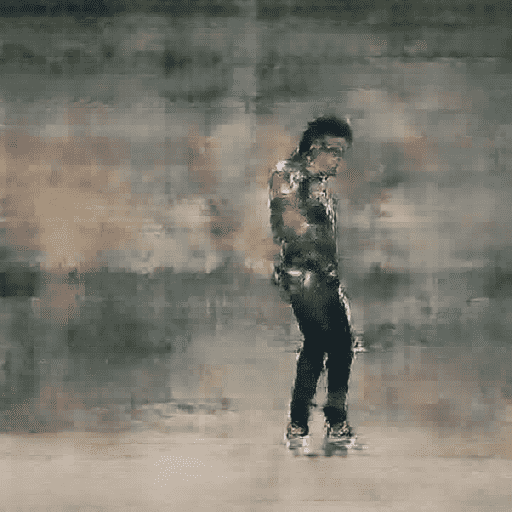} &
  \includegraphics[width=0.25\linewidth]{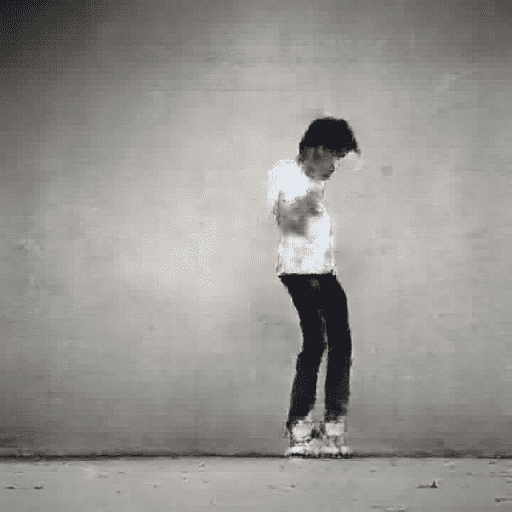} \\

  \end{tabular}
  }  
  \begin{minipage}{\textwidth}
    \caption[Compact Routing Example]{
  {\bf Visualizing Initializations.}\footnote{Video visualization is available at \url{https://youtu.be/zFoT8VcbwsU}}
  We visualize knowledge captured by meta-learning by running the \method{}-trained model (MT) without finetuning on a 
  test video. We fix $K = 5$, vary $T = [20, 40, 100, 200]$.  
  The pre-trained model (PT) generates an image from its training set.
  As we increase $T$, \method{} learns to {\em factor} pose and appearance, generating consistent background and clothing appearance regardless of the human pose. Figure \ref{fig:temp_coherence_ft} demonstrates that such factored representations are easier to fine-tune, 
  and result in more temporally stable generated videos.}
  \end{minipage}

  \label{fig:viz_init}
\end{figure} 
\end{document}